\documentclass[conference,compsoc]{IEEEtran}

\usepackage[utf8]{inputenc}
\UseRawInputEncoding 
\usepackage{verbatim}
\usepackage{url}
\usepackage{cite}
\usepackage{amsmath,amssymb,amsfonts}
\usepackage{algorithmic}
\usepackage{graphicx}
\usepackage{textcomp}
\usepackage{soul}
\usepackage{amsmath}

\DeclareMathOperator{\A}{\mathbf{A}}
\DeclareMathOperator{\Ll}{\mathbf{L}}

\DeclareMathOperator{\D}{\mathbf{D}}
\DeclareMathOperator{\Dn}{\mathbf{D^{-\frac{1}{2}}}}

\DeclareMathOperator{\I}{\mathbf{I}}

\DeclareMathOperator{\gt}{g_{\mathbf{\theta}}}

\DeclareMathOperator{\Uu}{\mathbf{U}}
\DeclareMathOperator{\UT}{\mathbf{U^{\intercal}}}
\DeclareMathOperator{\g}{\mathbf{g}}
\DeclareMathOperator{\LBD}{\mathbf{\Lambda}}

\DeclareMathOperator{\N}{\mathbf{N}}

\makeatletter
\newcommand{\linebreakand}{%
  \end{@IEEEauthorhalign}
  \hfill\mbox{}\par
  \mbox{}\hfill\begin{@IEEEauthorhalign}
}
\makeatother

\author{

\IEEEauthorblockN{Kourosh T. Baghaei\textsuperscript{\textsection}}
\IEEEauthorblockA{Department of Computer Science\\ George Mason University\\
kteimour@gmu.edu}
\and

\IEEEauthorblockN{Amirreza payandeh\textsuperscript{\textsection}}
\IEEEauthorblockA{Department of Computer Science\\ George Mason University\\
apayande@gmu.edu}
\and

\IEEEauthorblockN{Pooya Fayyazsanavi}
\IEEEauthorblockA{Department of Computer Science\\ George Mason University\\
pfayyazs@gmu.edu}
\linebreakand 
\IEEEauthorblockN{Shahram Rahimi}
\IEEEauthorblockA{Department of Computer Science\\ Mississippi State University\\
rahimi@cse.msstate.edu}
\and
\IEEEauthorblockN{Zhiqian Chen}
\IEEEauthorblockA{Department of Computer Science\\ Mississippi State University\\
zchen@cse.msstate.edu}
\and
\IEEEauthorblockN{Somayeh Bakhtiari Ramezani}
\IEEEauthorblockA{Department of Computer Science\\ Mississippi State University\\
sb3182@msstate.edu}

}

\title{Deep representation learning: Fundamentals, Perspectives, Applications, and Open Challenges}

\begin{document}
\maketitle


\begingroup\renewcommand\thefootnote{\textsection}
\footnotetext{These authors contributed equally to this work}
\endgroup

\begin{abstract}

Machine Learning algorithms have had a profound impact on the field of computer science over the past few decades. These algorithms' performance is greatly influenced by the representations that are derived from the data in the learning process. The representations learned in a successful learning process should be concise, discrete, meaningful, and able to be applied across a variety of tasks. A recent effort has been directed toward developing Deep Learning models, which have proven to be particularly effective at capturing high-dimensional, non-linear, and multi-modal characteristics. In this work, we discuss the principles and developments that have been made in the process of learning representations, and converting them into desirable applications. In addition, for each framework or model, the key issues and open challenges, as well as the advantages, are examined.

\end{abstract}



\IEEEpeerreviewmaketitle

\section{Introduction}
\label{sec:introduction}
In recent years, machine learning has shown promising capabilities in various fields of study and application. Representation learning as a core component in artificial intelligence attracts more and more scientists every day. This interest is mirrored in the increasing range of papers,  publications, and workshops on representation learning in international conferences and various influential journals.

Representation learning involves in detecting, extracting, encoding and decoding features from raw data, that may be used in learning tasks. In other words, representation learning is used to abstract features that best represent data. The algorithms developed for this task are referred to as Representation Learning\cite{lecun_deep_2015}. The performance of deep learning models, highly relies on the methods of representation of data. Therefore, the dramatic growth in deep learning has been accompanied by many significant developments within the underlying techniques in representation learning. Deep Learning lends its success to the architectures composed of multi-layered non-linear modules that each transform features to a higher level representation.

Learning representation aims to encode (embed) the raw input data into a real-valued vector of lower dimensions (embeddings), ideally disentangling the features that cause variation in the data distribution. It is expected that small differences in the input data or outliers would not affect most of these features. Hence, temporally or spatially close samples are expected to fall into very close locations in the space of representations. The deep representation learning methods have made a hierarchical structure of descriptive factors possible. In other words, the concepts represented in their higher layers describe the more abstract concepts. Ideally, high-level representation is one with simple and linearly correlated factors \cite{bengio_representation_2014}. Owing to the nature of feature extraction in representation learning, the representations can be shared and utilized across different tasks. Although achieving the characteristics mentioned above is challenging, the learned representation facilitates the discovery of latent patterns and trends in data for the learner, hence enhancing the learning of the multiple tasks \cite{bengio_representation_2014}. Based on the application, the raw input data can be of any type, for instance, texts, images, audio, video, etc. Given a particular task such as classification, segmentation, synthesis, and prediction, the main objective is to update the parameters of a neural network so it can represent the input data in a lower dimension.

 In the Image processing domain, representation learning can be employed in visualization \cite{hinton_reducing_2006}, regression\cite{Ge_2018_ECCV,mosadegh2020estimating,rs14194979}, interpretation of the predictions \cite{lipton_mythos_2017,savadjiev_demystification_2019,chen_deep_2019}, generating synthetic data \cite{nie_medical_2018}, finding and retrieving similar cases of images \cite{zin_content-based_2018,wan_deep_2014}, enhancing and denoising images \cite{wolterink_generative_2017, litjens_survey_2017}semantic segmentation and object detection \cite{long_fully_2015,miao_cnn_2016, soheyla_image_recognition_dl_2018}. Many problems in 2D image processing may also have arisen in volumetric image processing contexts such as 3D MRI\cite{mazurowski_deep_2019} and Point Cloud data \cite{pointcloud}captured by depth sensors  \cite{liu_deep_2019}.
 In the analysis of sequential data, representations may be transferred across domains to generate annotations and captions for images \cite{shin_learning_2016, hossain_comprehensive_2019,soheyla_image_caption_review_2019ASR}, and adding post-hoc interpretation to medical data analysis \cite{retainvis_2018}.
 
Natural language processing leverages representation learning approaches in various domains such as text classification\cite{nlp_text_classification}, question answering\cite{nlp_question_answering}, machine translation\cite{bahdanau2014neural,wu2016google,luong2015effective}, electronic health records\cite{esteva2019guide}, financial forecasting\cite{nlp_financial_survey}, etc. 
Historically, the beginning of NLP research started with rule-based methods. Later, access to the large amount of data made it possible to apply statistical learning methods to NLP tasks. By introducing deep learning approaches to NLP in 2012, neural network-based NLP has become the dominating approach in this field\cite{nlp_history}. In modern NLP, Word2vec\cite{mikolov2013}, and Glove\cite{pennington-etal-2014-glove}, can be viewed as two of the most advanced and well-known ways of representing words as vectors. After the 2017 breakthrough in the NLP community by transformers
\cite{vaswani2017attention}, advanced pre-trained models, in particular, BERT \cite{bert}
have become the center of attention and caused a stir in the
NLP community.

Linear factor models such as PCA and ICA are among the earliest methods of feature extraction in representation learning. Although these simple models may be extended to form more powerful models, this article concentrates more on deep models of representation. Readers might consult  \cite{goodfellow_deep_2016,bengio_representation_2014} for in depth discussion on these topics. The prevalent approaches to deep representation learning are explained in the following sections.

\section{Multi Layer Perceptron}
\label{sec:anns}
\begin{figure}[hbtp]
    \centering
    \includegraphics[width=0.75\linewidth]{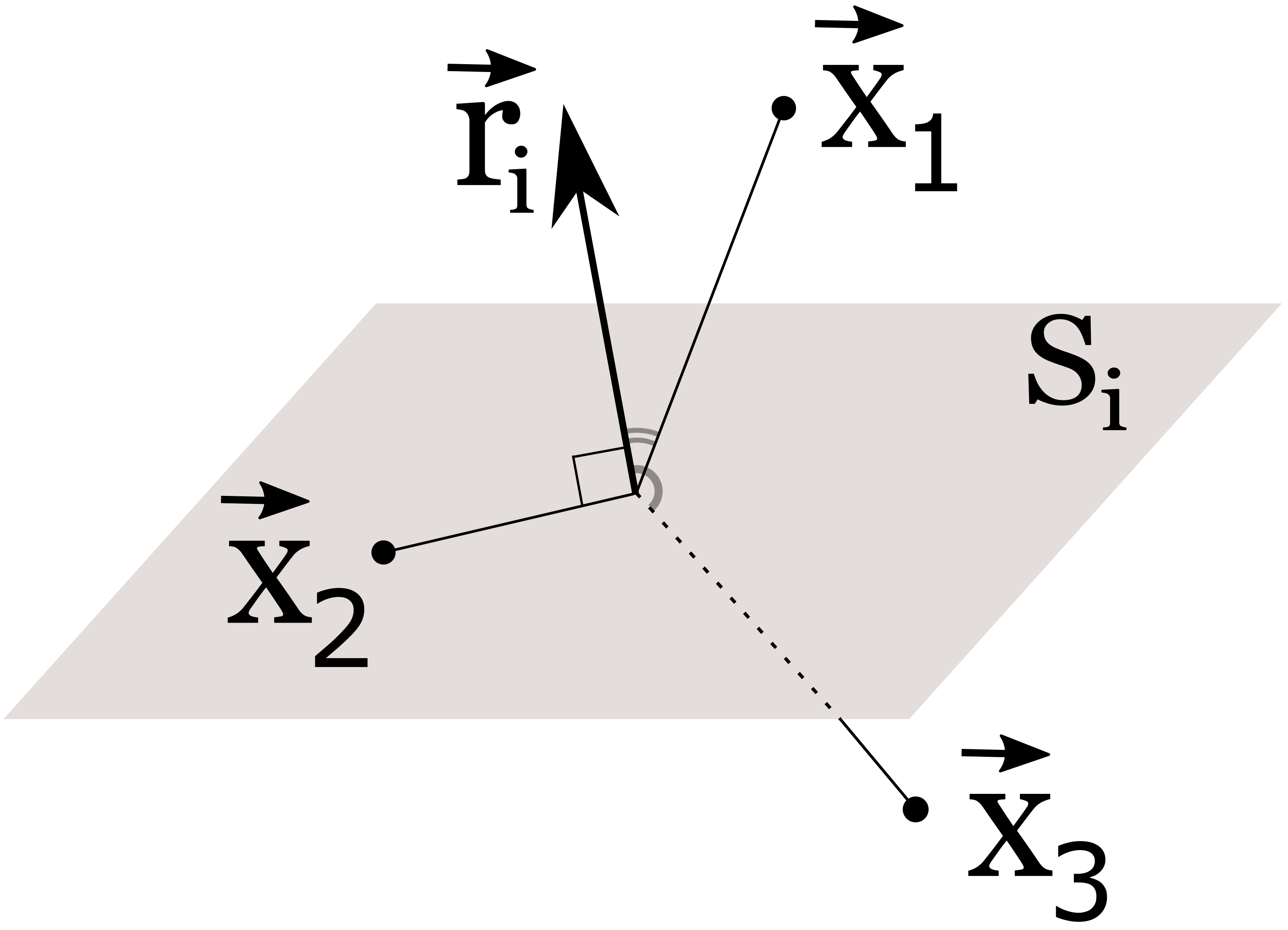}
    
    \caption{An intuitive representation of how the weights of a linear layer transform from the input space to the output space. $\vec{r}_i$, the $i$-th row of the weight matrix of the linear layer may be considered as the normal vector of surface $S_i$, that may affect different input data points in different ways: a) Points above the surface: $\vec{r}_i \cdot \vec{x}_1 >  0 $. b) Points residing on the surface: $\vec{r}_i \cdot \vec{x}_1 = 0 $. c) Points that are posed below the surface: $\vec{r}_i \cdot \vec{x}_1 < 0$. The result of the product may also be passed through an activation function to add non-linearity.}

    \label{fig:mlp}
\end{figure}

A multi layer perceptron or feedforward neural network is a stack of multiple layers. Each layer, consists of one linear transformation and one non-linear activation function. Given an input vector $\vec{x} \in \mathbb{R}^n$ and weight matrix $W \in \mathbb{R}^{n\times m}$, transformed vector $\vec{y} \in \mathbb{R}^m$ can be calculated as:

\begin{equation}
    \label{eq:lin_trans}
    \vec{y} = W^{T}\vec{x}
\end{equation}
The weight matrix $W$ in Eq. \ref{eq:lin_trans} consists of $m$ rows $\vec{r}_i \in \mathbb{R}^m$ (where $ 1\leq i \leq m$). As depicted in Fig. \ref{fig:mlp}, each row $\vec{r}_i$ can be thought of as a vector perpendicular to a surface $S_i$ in hyperspace that passes through the origin. Surface $S_i$ divides the $n$-dimensional space into 3 sub-spaces: three sets of points residing on the surface and the two sides of it. Each $y_i$ in vector $\vec{y}=(y_1,y_2,...,y_m)$, is calculated by the dot product of row $\vec{r}_i$ and the input vector $\vec{x}$. Depending on the relative positions of the point $\vec{x}$ and surface $S_i$, the value of $\vec{y}$ along the $i$-th dimension may be positive, negative, or zero. A bias number $b_i$ can also be employed to further control the value of $\vec{y}$. Essentially, the parameters of the weight ($\vec{r}_i$) and bias ($\vec{b}$) vectors decide on how the features of the input vector $\vec{x}$ affect $\vec{y}$ along the $i$-th dimension in the target space of $m$ dimensions. The training process, updates these weights and biases so that they can fit the input data to their corresponding target values. Thus, the network learns how to distinguish or generate certain similarities and patterns among the features of the input data.
Each of the parameters of $\vec{y}$ are passed to an activation function in order to add non-linearity to the output. In a similar way, an extra layer can be utilized to capture the patterns and similarities in the output vectors of the previous layer. Hence, extracting more complex characteristics in the data. Adding extra layers may increase the capability of a network in learning representations in exchange for its computational complexity.

\section{Generative Models}
\label{sec:gen_models}
Generative models, are unsupervised methods that aim to learn and approximate the distribution function by which the samples of a given unlabeled dataset are generated. Having the approximate generator function learned, a model would be able to randomly generate samples that are not present in the actual dataset, though resemble them \cite{oussidi_deep_2018}. This functionality has been applied in numerous different domains such as machine translation (MT), question answering and natural language generation\cite{torfi2020natural}. Furthermore, generative models are used in image processing tasks such as: data augmentation \cite{yi_generative_2019}, image denoising \cite{jiang_denoising_2018}, reconstruction \cite{hammernik_learning_2018} and super-resolution \cite{lyu_super-resolution_2018}. Generative models may be grouped into two categories: energy-based and function-based models\cite{oussidi_deep_2018}. Boltzmann Machines(BM), Restricted Boltzmann Machines(RBM) and Deep Belief Networks(DBN) are the examples of energy-based models \cite{hinton_fast_2006}. On the other hand, the Auto-Encoder \cite{rumelhart_learning_1986,hinton_reducing_2006} and its variants and Generative Adversarial Networks \cite{goodfellow_generative_2014} belong to the function-based group. \par

\subsection{Boltzmann Machines}

\textbf{Boltzmann Machine} is an energy-based model primarily introduced for learning arbitrary probability distribution over binary vectors \cite{goodfellow_deep_2016}. Although continuous variations of it are later proposed \cite{lecun2015deep}.
Given a $d$ dimensional binary vector $x \in \{0,1\}^d$ as input, the joint probability distribution is defined as:
 
\begin{equation}
    P(x) = \frac{exp(-E(x))}{Z}
\label{eq:boltz_bin}
\end{equation}

Where Z is normalization parameter defined as:

\begin{equation}
    Z = \sum_{x}^{}{exp(-E(x))}
\end{equation}

So that the P(x) form a probability density. And E(x) in the Eq. \ref{eq:boltz_bin} is the energy function defined as:

\begin{equation}
    E(x)=-(x^{T}Wx + b^{T}x)
\end{equation}

The training process involves in maximizing the likelihood and minimizing the energy function. 
The learning procedure in Boltzmann machines similar to biological neurons. Meaning that a connection is strengthened if both neurons at its ends are excited together and weakened otherwise.
\par

\textbf{Restricted Boltzmann Machine} limits the connection among graph's nodes to only links between the visible and hidden neurons. Henceforth, there are connections among neither hidden neurons nor visible ones. The vector of nodes $x$, can be represented as two subsets: visible nodes $v$ and hidden nodes $h$.
So the energy function will be:
 
 \begin{equation}
    \text{E}(v,h)= -b^{T}v - c^{T}h - v^{T}Wh
\end{equation}

In which $b$ and $c$ represent bias weights and the matrix $W$ represents connections' weights. 
 
The partition function for RBM is:
 
\begin{equation}
    Z = \sum_{v}^{}\sum_{h}^{}{exp(-E(v,h))}
\end{equation}
 
RBMs are probabilistic graphs and the building blocks of Deep Belief Networks (DBN). Since partition function $Z$ is intractable, RBMs should be trained with methods such as Contrastive Divergence \cite{CarreiraPerpin_2005_On_Contrastive_Divergence} and Score Matching \cite{goodfellow_deep_2016}.
 
\textbf{Deep Belief Network} consists of several stacked RBMs. A DBN with only one hidden layer is an RBM. In order to train a DBN, first, an RBM is trainned by likelihood maximization or contrastive divergence \cite{goodfellow_deep_2016}. Then, another RBM is trained to model the distribution of the first one. More RBMs can be added to this stack in order to form a deep network of RBMs. Adding up layers increases the variational lower bound of the log-likelihood of the data.
The deepest layer of DBNs has undirected connections, in contrast to other deep neural network architectures\cite{hinton_fast_2006}; However, the term DBN is incorrectly used to refer to any neural network.

\textbf{Other Variants} of Boltzmann machines have been proposed such as Deep Boltzmann Machines (DBM) \cite{salakhutdinov_deep_boltzmann_2009a}, Spike and Slab Restricted Boltzmann Machines (ssRBM) \cite{courville_spike_slab_rbm_2011},  Convolutional Boltzmann Machines \cite{lee_conv_rbm_2009}. However, other generative models such as variational auto encoders and generative adversarial networks have proved as viable substitutes for variations and derivations of Boltzmann machines.

\subsection{Auto-Encoders}
\label{subsec:autoenc}
Autoencoder-based models are considered to be some of the most robust unsupervised learning models for extracting effective and discriminating features from a large unlabeled dataset. The general architecture of an auto-encoder consists of two components: \par
\textbf{Encoder:} function $f$ which aims to transform the inputs $x$ to a latent variable $h$ in lower dimensions. \par
\textbf{Decoder:} function $g$  reconstructs the input $\hat{x}$, given the latent variable $h$.
The training process involves updating the weights of the encoder and decoder networks according to the loss function of the reconstruction:
\begin{equation}
    \mathcal{L}(x,\hat{x}) = \mathcal{L}\Big(x,g\big(f(x)\big)\Big)
\end{equation}


Many variants of auto-encoders have been proposed in the literature; however, they can be categorized in four major groups \cite{oussidi_deep_2018}. \par

\textbf{Undercomplete Autoencoder:}
In order to make the autoencoder learn the distributions from the data, the latent variables should have lower dimensions than the input data. Otherwise, the network would fail to learn any useful features from the data. This type of autoencoder is known as \textit{undercomplete autoencoder}\cite{goodfellow_deep_2016}. 

\begin{figure}[!hbtp]
\centering
    \includegraphics[width=0.95\linewidth]{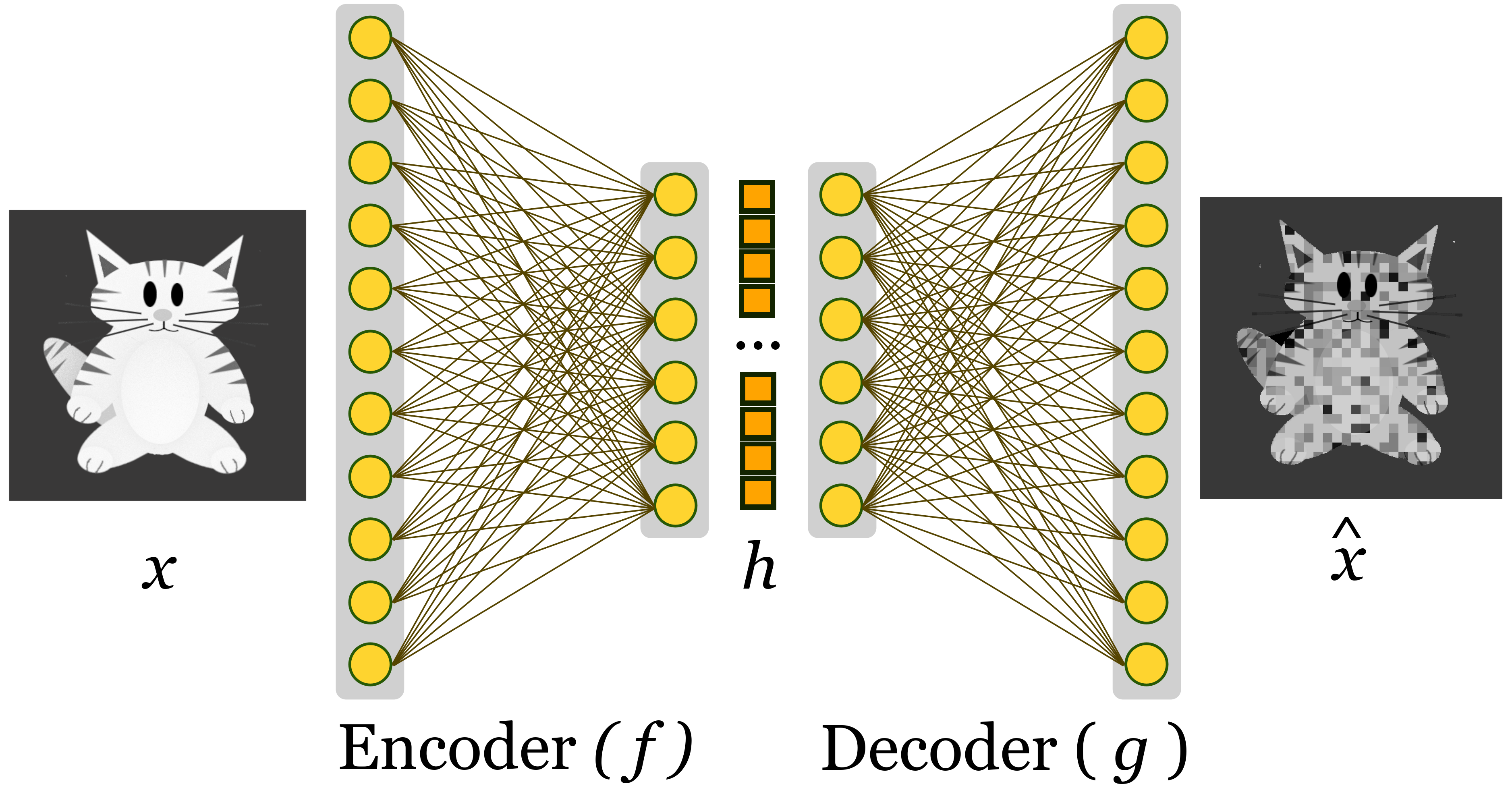}
    \caption{General architecture of an auto-encoder. The encoder transforms input $x$ into the latent vector $h$: $h = f(x)$. The decoder reconstructs the input from $h$: $\hat{x} = g(h)$.}
    \label{fig:autoenc}
\end{figure}

\begin{figure}[hbtp]
    \centering
    \includegraphics[width=1.\linewidth]{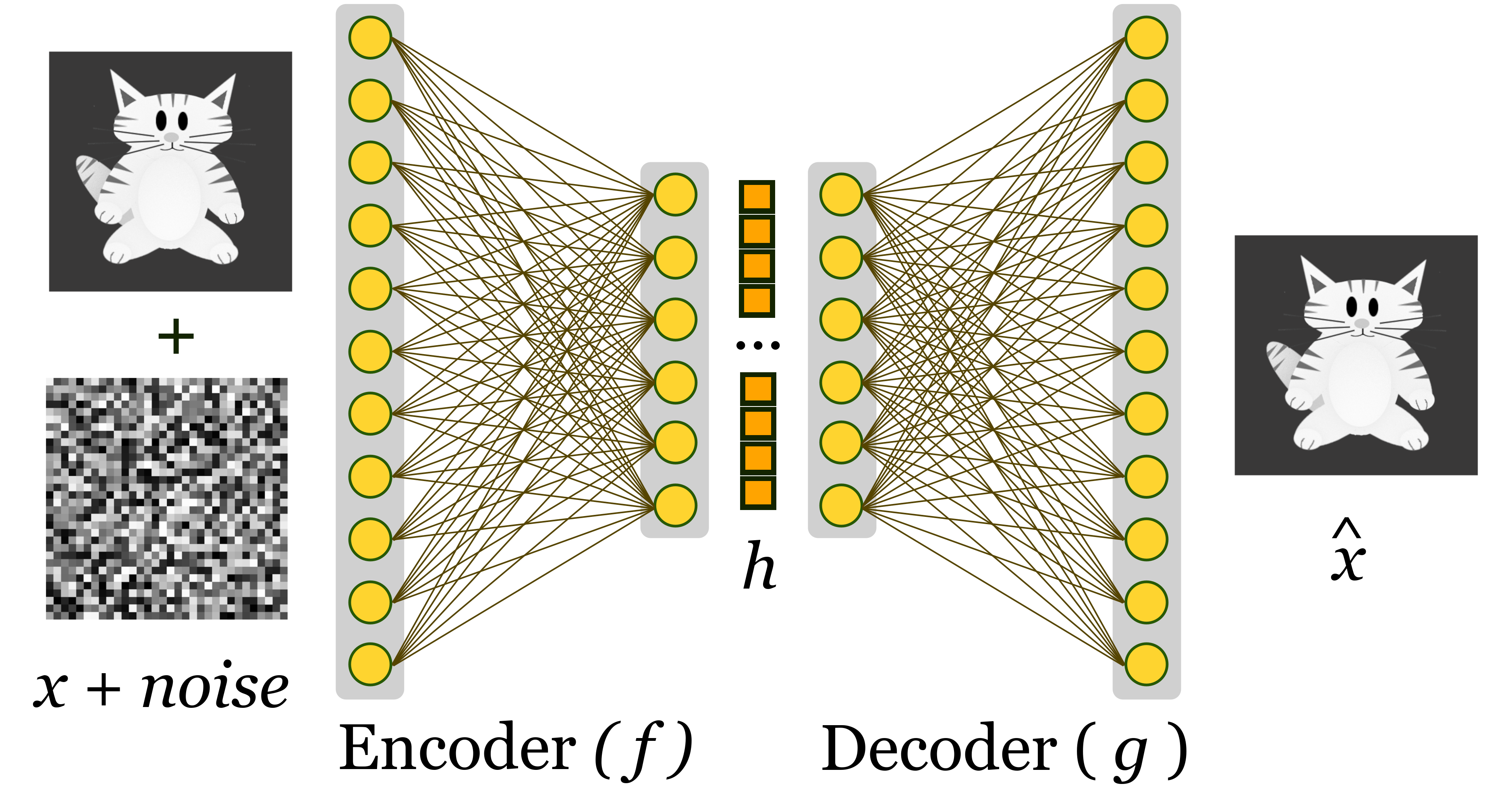}
    \caption{General architecture of a denoising auto-encoder (DAE). Adding noise to the input during the training process, results in more robust learning of the features. Hence, increasing the generalization ability. }
    \label{fig:denoising-ae}
\end{figure}

\textbf{Denoising Autoencoder (DAE):}
Denoising Autoencoder corrupts the data by adding stochastic noise reconstructs it back into intact data.  Hence, it is called \textit{denoising autoencoder}. As depicted in Fig. \ref{fig:denoising-ae}, the added noise to the input is the only difference of this method to the traditional autoencoders. This approach results in better feature extraction and better  generalization in classification tasks \cite{vincent_stacked_2010}.  Also, Several DAEs can be trained locally by adding noise to their inputs and stacking consecutively to form a deep architecture called Stacked DAE,  with higher representation capabilities.
 
\textbf{Sparse Autoencoders (SAE)}:
Sparse Representation refers to the technique of decomposing a data set into a set of overcomplete vectors where only a small subset of those vectors combine to describe the data. The overcompleteness of representation can lead to more expressive basis vectors which can better capture more complex structures. The sparsity puts an additional constraint on the number of basis vectors present for decomposing data to basis vectors.
Sparse representation can be formulated as disentangling of an input signal into a linear combination of its latent features \cite{bengio_learning_2009}. The loss function of a sparse autoencoder includes an additional sparsity constraint ($\Omega(h)$) on the latent variables \cite{goodfellow_deep_2016}:
\begin{equation}
\label{eq:sparse_enc}
    \mathcal{L} = \mathcal{L}\big(x,g(f(x))\big) + \Omega(h)
\end{equation}
Thus making the autoencoder to extract features from the data and represent them in sparse vectors and matrices \cite{giron-sierra_sparse_2017}.

\textbf{Variational Autoencoder (VAE):}
Although this type of autoencoder has the same components as the traditional autoencoder (Fig. \ref{fig:autoenc}), its training process is based on variational inference \cite{blei_variational_2017}. Just as the traditional autoencoder, the encoder function $f$ is trained to map the input data to the latent variables $z$ and the decoder function $g$ is trained to map the latent variables $z$ to the input data. However, for this autoencoder to work, the latent variable $z$ is assumed to be Guassian\footnote{ Depending on the type of data, this can also be Bernouli.}. By choosing this representation, we gain significant control over how the latent distribution should be modeled, resulting in a smoother and more continuous latent space. And the loss function for this training consists of two parameters: First,  \textit{Kullback-Leibler(KL) divergence}\cite{kullback1951} of the output of the encoder $f$ and Guassian distribution; Thus forcing the encoder to map the input data to the Gaussian distribution in the latent space. Second, the reconstruction loss: \cite{kingma_auto-encoding_2014}:
\begin{equation}
\label{eq:var_ae}
    \mathcal{L} = \mathcal{D}\big(KL(f\| \mathcal{N}(0,I))\big) +  \mathcal{L}\big(x,g(f(x))\big)
\end{equation}
Variational inference is discussed in more details in section \ref{subsubsec:variational}.

\textbf{Contractive Autoencoder (CAE):}
Making the features in the activation layer invariant with respect to the small perturbations in input, was the main goal in proposing this variant of autoencoder \cite{rifai_contractive_autoenc_2011}. The basic autoencoder may be converted to a contractive autoencoder by adding the following regularization to its loss function:

\begin{equation}
\label{eq:cae}
    \| J_f(x) \|_{F}^{2} = \sum_{i}\sum_{j}\Big( \frac{\partial h_j(x)}{\partial x_i} \Big)^2
\end{equation}

Where $f : \mathbb{R}^m \rightarrow \mathbb{R}^n$ is a non-linear mapping function from input space $x \in \mathbb{R}^m$ to the hidden layer $h \in \mathbb{R}^n$.
The regularization term is the squared value of the first-order partial derivatives of the hidden values with respect to input values. By penalizing the first derivative of the encoding function, the derivative is forced to maintain lower values. In this way, the encoding function may learn a flatter representation. As a result, the encoding function may become more robust or invariant to small perturbations in the input.

The loss function of the contractive autoencoder may be written as:
\begin{equation}
\label{eq:cae_loss}
    \mathcal{L}_{CAE} = \sum_{x \in X}\bigg(\mathcal{L}_R\big(x,g(f(x))\big) + \lambda\| J_f(x) \|_{F}^{2} \bigg)
\end{equation}

Where $X$ is the dataset of training samples, $\mathcal{L}_R$ denotes the reconstruction loss, and $\lambda \in \mathbb{R}$ controls the effect of contractive loss. The input points get closer in distance when mapped to the hidden state i.e. they are \textit{contracted}. This contraction can be thought as the reason behind robustness in features.

\subsection{Generative Adversarial Networks (GAN)}
\label{subsec:gans}

\begin{figure}[htb]
\centerline{\includegraphics[width=3.3in]{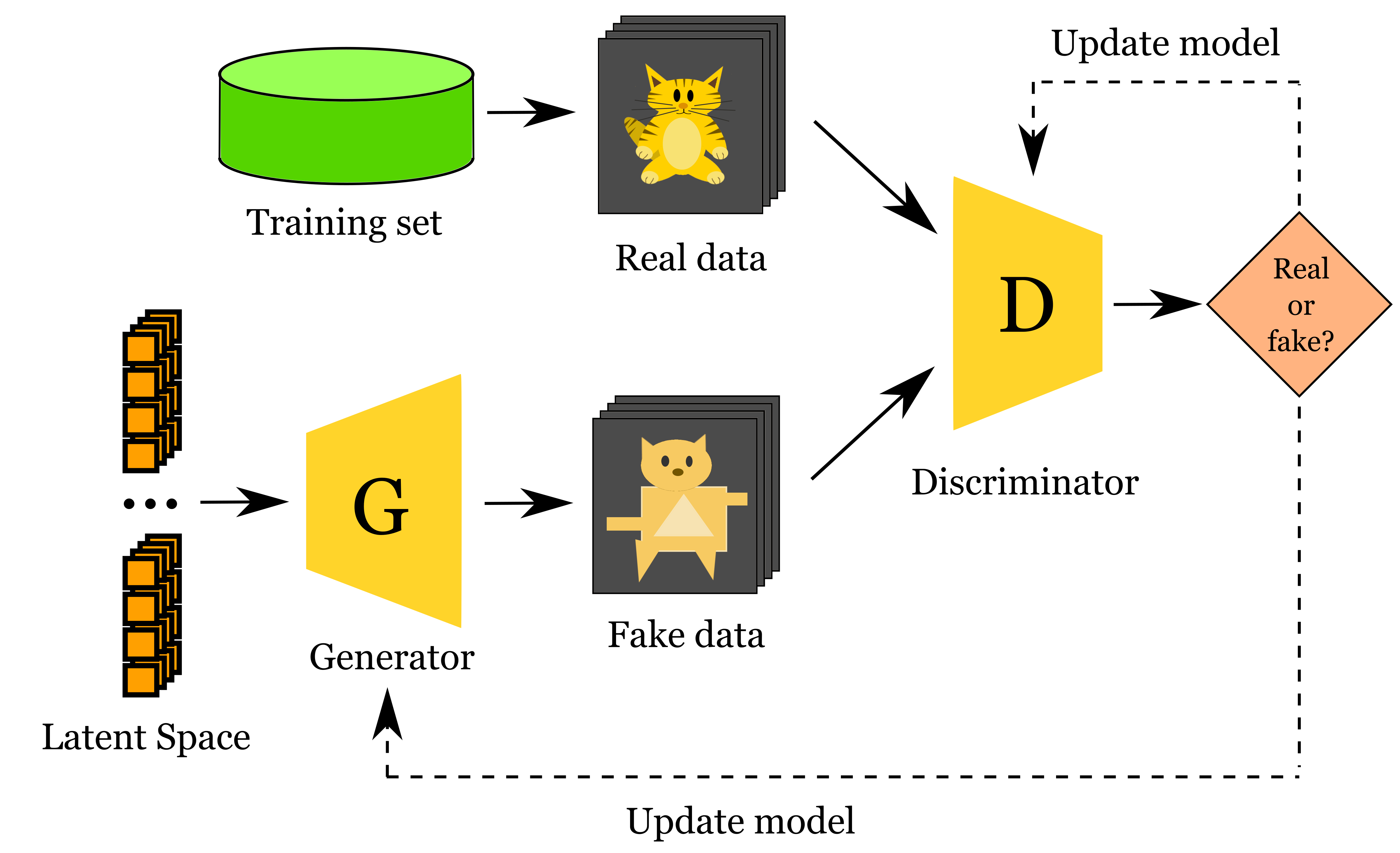}}
\caption{Generative Adversarial Network.}
\label{fig:gan}
\end{figure}

Although both Autoencoders and GANs are generative models, their learning mechanism is different. Autoencoders are trained to learn hidden representations, where GANs are designed to generate new data. The most prevalent generative model utilized in many applications is the Generative Adversarial Network (GAN) architecture \cite{goodfellow_generative_2014}. As depicted in Fig. \ref{fig:gan}, it resembles a two-player minimax game where two functions known as the generator $\mathcal{G}$ and the discriminator $\mathcal{D}$ are trained as opponents. The $\mathcal{G}$ function tries to generate fake samples as similar as possible to the real input data from a noise variable $z$, and the $\mathcal{D}$ function aims to discriminate the fake and real data apart. The minimax game can be described with the following objective function:
\begin{equation}
\label{eq:gan_minmax}
\begin{split}
    \underset{\mathcal{G}}{min}\ \underset{\mathcal{D}}{max}\ V(\mathcal{D},\mathcal{G})
    &= \mathbb{E}_{x\sim p_{data}(x)}[log \mathcal{D}(x)]\ + \\
    & \mathbb{E}_{z\sim p_{z}(z)}[log \big( 1 - \mathcal{D}(\mathcal{G}(z))\big)]
\end{split}
\end{equation}
where $x \sim p_{data}$ denotes the real data sample $x$ with its distribution $p_{data}$. And $\mathcal{D}(x)$ represents the class label that the discriminator $\mathcal{D}$ assigns to the input sample $x$. For the noise variable $z$ a prior is assumed as $z \sim p_z(z)$. \par
The success of CNNs in image analysis and the capabilities that GANs provide, has made generative CNNs possible  \cite{radford_unsupervised_2016}.
Numerous extensions to the original GAN have been proposed so far \cite{creswell_generative_2018} such as interpretable representation learning by information maximizing (InfoGAN) that forces the model to disentangle and represent features of images in certain elements of the latent vector \cite{chen_infogan_2016}. Or Cycle-Consistent GAN (CycleGAN) that learns characteristics of an image dataset and translates them into another image dataset without any dataset of paired images \cite{zhu_unpaired_2017}. An inherent limitation of the original GAN is that it does not have any control over its output. Conditional Generative Adversarial Nets \cite{https://doi.org/10.48550/arxiv.1411.1784} incorporate auxiliary inputs such as class labels into their model to generate the desired output.

\subsection{Applications}
Generative neural networks empower countless deep architectures across various disciplines. Natural Language Processing \cite{Kim2018BridgenetsST,hinton2012deep}, generating sentences \cite{Chen2019AdversarialSF_text,bowman_2015_gen_sentences, chen2019reinforcement, soheyla_image_caption_review_2019ASR}. Generative models can empower deep architectures to achieve state-of-the-art performance in numerous problems in image processing. Denoising of 3D magnetic images \cite{ran_denoising_2019}, unsupervised image generation \cite{munir_cancer_2019} , image to image translation  \cite{zhu_unpaired_2017,zhang_translating_2018}, image cross-modality synthesis \cite{li_2019_ct_mri,jin_deep_2019}, data augmentation and anonymization \cite{shin_medical_2018}, segmentation \cite{kamnitsas_efficient_2017,alansary_fast_2016}, super resolution \cite{creswell_generative_2018,wang_high-resolution_2018,bahrami_2016_7T,qu_2020_7T}, and video analysis \cite{huang_introduction_2018}. Moreover, generative neural networks and their derivatives utilize deep reinforcement learning algorithms \cite{chen_deep_rl_2019,wu_deep_rl_2018_object_detection}, and are employed in analysis of graph data  \cite{Li_2019_CVPR,li2018deeper}.

\section{Graph Neural Networks}
\label{sec:gnn}
The success of deep learning has been widely demonstrated in numerous applications during the last decade \cite{lecun2015deep, redmon2016you,ren2015faster,hinton2012deep,wu2016google,luong2015effective}. Among the recent study of deep learning, graph neural networks (GNNs) is treated as one significant progress towards effective information analysis in the non-Euclidean geometry of the data. GNN has covered many real-world scenarios such as gene data on biological regulatory networks \cite{davidson2002genomic}, telecommunication networks \cite{drew2008diagnosing}, social networks \cite{lazer2009life}, transportation networks \cite{bell1997transportation}, spread of epidemic disease \cite{newman2002spread}, and brain's neuronal networks \cite{marx2012high}. 
However, the non-Euclidean structure of the graph cannot be handled by previous deep learning, such as ConvNet. This is because the graph is irregular since each node has a different size of neighbors, which conflicts with the fixed size of the kernel in ConvNet. Borrowing the power of the success of deep learning, many graph neural networks are proposed to model the intrinsic complexity of non-Euclidean graph \cite{bruna2014spectral,kipf2016semi,defferrard2016convolutional,hamilton2017inductive,atwood2016diffusion,velivckovic2017graph}, attracting increasing interest in the machine learning community. Among all the GNNs, graph convolutional network (GCN) is the most popular model, and it nicely integrates the graph topology with node properties to generate fused representations. There exist a huge number of different mechanisms behind GCNs, such as random walk, Page Rank, attention model, low-pass filter, message passing, etc. In the perspective of graph convolution, those methods can be categorized into several coarse-grained groups \cite{monti2017geometric,hamilton2017representation,zhang2018deep,zhou2018graph,wu2019comprehensive} such as spectral \cite{bruna2014spectral,kipf2016semi,defferrard2016convolutional} and spatial domain \cite{hamilton2017inductive,atwood2016diffusion,velivckovic2017graph}. 

\subsection{Basics of GNN}
Graph convolution originates from spectral graph theory which is the study of the properties of a graph in relationship to the eigenvalues, and eigenvectors of associated graph matrices \cite{chung1997spectral,grone1990laplacian,das2004laplacian}. The spectral convolution methods \cite{bruna2013spectral,defferrard2016convolutional,kipf2017semi,bronstein2017geometric} are the major algorithm designed as the graph convolution methods, and it is based on the graph Fourier transform \cite{shuman2013emerging, shuman2016vertex}. 
GCN focus processing graph signals defined on undirected graphs $\mathcal{G} = (\mathcal{V}, \mathcal{E}, \mathcal{W})$, where $\mathcal{V}$ is a set of n vertexes, $\mathcal{E}$ represents edges and
$\mathcal{W} = [w_{ij}] \in \{0,1\}^{n\times n}$ is an unweighted adjacency matrix. A signal $x : \mathcal{V} \rightarrow \mathbb{R}$ defined on the nodes may be regarded as a vector $x \in \mathbb{R}^{n}$.
Combinatorial graph Laplacian\cite{chung1997spectral} is defined as $\Ll= D-\mathcal{W} \in \mathbb{R}^{n\times n}$ where $D$ is degree matrix.
As $\Ll$ is a real symmetric positive semidefinite matrix, it has a
complete set of orthonormal eigenvectors and their associated ordered real nonnegative eigenvalues identified as the frequencies of the graph. The Laplacian is diagonalized by the Fourier basis $\UT$: $\Ll = \Uu \Lambda \UT$ where $\Lambda$ is  the diagonal matrix whose diagonal elements are the corresponding eigenvalues, i.e., ${\displaystyle \Lambda _{ii}=\lambda _{i}}$. The graph Fourier transform of a signal $x\in \mathbb{R}^{n}$ is defined as $\hat{x}=\UT x \in \mathbb{R}^{n}$ and its inverse as $x=\Uu \hat{x}$\cite{shuman2013emerging, shuman2016vertex, zhu2012approximating}. To enable the formulation of fundamental operations such as filtering in the vertex domain, the convolution operator on graph is defined in the Fourier domain such that $f_{1}*f_{2}=\Uu \left[\left(\UT f_{1} \right) \odot \left(\UT f_{2}\right)\right]$, where $\odot$ is the element-wise product, and $f_{1}/f_{2}$ are two signals defined on vertex domain. It follows that a vertex
signal $f_{2}=x$ is filtered by spectral signal $\hat{f_{1}}=\UT f_{1}=\g$ as:
\begin{equation*}
\g * x = \Uu \left[\g(\LBD)\odot \left(\UT f_{2}\right)\right] = \Uu \g(\LBD) \UT x.
\end{equation*}

\begin{figure}[!hbtp]
\centering
    \includegraphics[width=1.\linewidth]{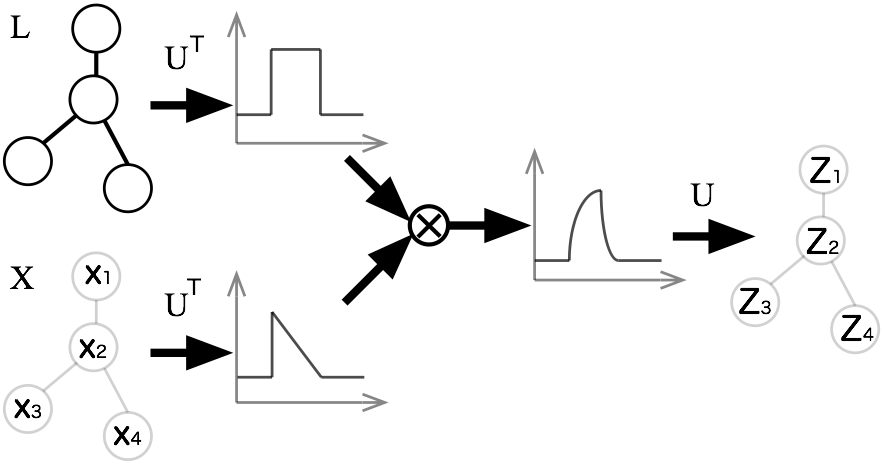}
    \caption{Illustration of Graph Convolution}
    \label{fig:gcn}
\end{figure}

Note that a real symmetric matrix $\Ll$ can be decomposed as $\Ll=\Uu \LBD \Uu^{-1} =  \Uu \LBD \UT$ since $\Uu^{-1}=\UT$ . D. K. Hammond et al. and Defferrard et al.\cite{hammond2011wavelets, defferrard2016convolutional} apply polynomial approximation on spectral filter $\g$ so that:
\begin{align*}
& \g * x = \Uu \g (\LBD) \UT x &&\\
\approx&\Uu \sum_{k}^{}\theta_{k} T_{k}(\tilde{\LBD}) \UT x && {(\tilde{\LBD}=\frac{2}{\lambda_{max}}\LBD-\I_{\N})}\\
=&\sum_{k}^{}\theta_{k} T_{k}(\tilde{\Ll}) x &&{(\Uu\LBD^{k}\UT=(\Uu\LBD\UT)^{k})}\\
\end{align*} Kipf et al.\cite{kipf2017semi} simplifies it by applying multiple tricks:
\begin{align*}
&\g * x &&\\
\approx& \theta_{0}\I_{\N}x+\theta_{1}\tilde\Ll x &&({\scriptstyle\text{expand to 1st order)}}\\
=&\theta_{0}\I_{\N}x+\theta_{1}(\frac{2}{\lambda_{max}}\Ll-\I_{\N})) x &&{\scriptstyle(\tilde\Ll=\frac{2}{\lambda_{max}}\Ll-\I_{\N}))} \\
=&\theta_{0}\I_{\N}x+\theta_{1}(\Ll-\I_{\N})) x &&{\scriptstyle(\lambda_{max}=2)} \\
=&\theta_{0}\I_{\N}x-\theta_{1} \Dn\A\Dn x &&{\scriptstyle(\Ll=\I_{\N}-\Dn\A\Dn)} \\
=&\theta_{0}(\I_{\N} + \Dn\A\Dn) x &&{\scriptstyle(\theta_{0}=-\theta_{1})} \\
=&\theta_{0}(\tilde\D^{-\frac{1}{2}}\tilde\A\tilde\D^{-\frac{1}{2}}) x &&{\scriptstyle(\text{renormalization}: \tilde\A=\A+\I_{\N},}\\
&&&{\scriptstyle \tilde\D_{ii}=\sum_{j} \A_{ij})}.
\end{align*}Rewriting the above GCN in matrix form: $\gt*X \approx(\tilde\D^{-\frac{1}{2}}\tilde\A\tilde\D^{-\frac{1}{2}}) X\Theta$, it leads to \textit{symmetric normalized Laplacian} with raw feature. GCN has been analyzed GCN in \cite{DBLP:journals/corr/abs-1801-07606} using smoothing Laplacian\cite{taubin1995signal}, and the updated features ($y$) equals to the smoothing Laplacian, i.e., the weighted sum of itself ($x_{i}$) and its neighbors ($x_{j}$):
$y= (1-\gamma) x_{i} + \gamma \sum_{j}\frac{\tilde a_{ij}}{d_{i}}x_{j} 
= x_{i} - \gamma (x_{i}-\sum_{j}\frac{\tilde a_{ij}}{d_{i}}x_{j}),
$where $\gamma$ is a weight parameter between the current vertex $x_{i}$ and the features of its neighbors $x_{j}$, $d_{i}$ is degree of $x_{i}$, and $y$ is the smoothed Laplacian. Rewriting in matrix form, the smoothing Laplacian is:
\begin{align*}
Y&= x -\gamma \tilde\D^{-1}\tilde\Ll x&& \\
&= (\I_{\N}-\tilde\D^{-1}\tilde\Ll )x && (\gamma=1)\\
&= (\I_{\N}-\tilde\D^{-1}(\tilde\D-\tilde\A))x && (\tilde\Ll=\tilde\D-\tilde\A)\\
&= \tilde\D^{-1}\tilde\A x.&& \label{eq:smooth_laplacian}
\end{align*} The above formula is \textit{random walk normalized Laplacian} as a counterpart of \textit{symmetric normalized Laplacian}. Therefore, GCN can be treated as a first-order Laplacian smoothing which averages neighbors of each vertex.

\subsection{Taxonomy of GNN}
As many surveys on GNN state \cite{zhou2018graph,zhang2018deep,wu2019comprehensive,bronstein2017geometric,hamilton2017representation,chen2020bridging}, GCNs can be classified into two major categories based on the operation type. Therefore, we introduce a taxonomy of GNN in the following two perspectives.

\textbf{Spectral-based GNN:}
This group of GCN highly relies on spectral graph analysis and approximation theory. 
Spectral-based GNN models analyzes the weight-adjusting function (i.e., filter function) on eigenvalues of graph matrices, which corresponds to adjusting the weights assigned to frequency components (eigenvectors). Many of Spectral-based GNN models are equivalent to low-pass filters \cite{Li_2019_CVPR}. Based on the type of filter function, there are linear filtering \cite{kipf2016semi,hamilton2017inductive,xu2018how}, polynomial filtering \cite{hammond2011wavelets,perozzi2014deepwalk,atwood2016diffusion,grover2016node2vec,wu2019simplifying}, and rational filtering \cite{Li_2019_CVPR,bianchi2019graph,chen2018rational,klicpera2018predict}. Beyond that, \cite{li2022g} adaptively learns the center of spectral filter. Closely, \cite{fu2022p} proposed a high-low-pass filter based on p-Laplacian. \cite{yang2022new,kenlay2021interpretable,wang2022powerful} revisit the spectral graph convolutional filter and make theoretical analyze. Optionally, one can choose graph wavelet to model spectrum of each node \cite{meng2022earlya,meng2022earlyb,tang2022rethinking,zheng2021framelets}.

\textbf{Spatial-based GNN:}
Nowadays, there are more emerging GNNs using spatial operations. Based on the spatial operation, they can be categorized into three groups: local aggregation which only combine direct neighbors \cite{xu2018how,xu2018powerful,gilmer2017neural,hamilton2017inductive,velivckovic2017graph}, higher order aggregation which involves second order or higher orders of neighbors  \cite{atwood2016diffusion,hammond2011wavelets,defferrard2016convolutional,wu2019simplifying,tang2015line,grover2016node2vec}, and dual-directional aggregation that propagates information in both forward and backward directions \cite{chen2018rational,klicpera2018predict,Li_2019_CVPR,7131465,7581108,levie2018cayleynets,bianchi2019graph}.

\subsection{Applications}
Graph neural networks have been applied in numerous domains such as physics, chemistry, biology, computer vision, natural language processing, intelligent transportation, social networks \cite{zhou2018graph,zhang2018deep,wu2019comprehensive,bronstein2017geometric,hamilton2017representation}.
To model physical objects, DeepMind \cite{battaglia2018relational} provides a toolkit to generalize the operations on graphs, including manipulating structured knowledge and producing structured behaviors, and \cite{sanchezgonzalez2020learning} simulates fluids, rigid solids, and deformable materials. 
Treating chemical structure as a graph \cite{duvenaud2015convolutional,kearnes2016molecular,wang2022deep} represent molecular structure, and \cite{zitnik2018modeling,fout2017protein} model protein interfaces. Further, \cite{dai2019retrosynthesis} predict the chemical reaction and retrosynthesis. In computer vision, question-specific interactions are modeled as graphs in visual question answering \cite{norcliffe2018learning,narasimhan2018out}. Similar to physics applications, human interaction with humans could be represented by their connections \cite{qi2018learning,xu2019spatial,hu2018relation,gu2018learning,xu2019spatial}.
\cite{yao2019graph} model the relationship among word and document as a graph, while
 \cite{zhang2018graph,vashishth2018incorporating} characterize the syntactic relations as a dependency tree.
Predicting traffic flow is a fundamental problem in urban computing, and transportation network can be modeled as a spatiotemporal graph \cite{yao2018deep,zhang2018gaan,li2017diffusion,yu2017spatio}. 
Functional MRI (fMRI) is a graph data where brain regions are connected by functional correlation \cite{10.3389/fnsys.2010.00016,wang2017structural}. \cite{nandakumar2019novel} employs a graph convolutional network to localize eloquent cortex in brain tumor patients, \cite{arya2020fusing} integrates structural and functional MRIs using Graph Convolutional Networks to do Autism Classification, and \cite{craig2018deep} applies graph convolutional networks to classify mental imagery states of healthy subjects by only using functional connectivity.
To go beyond rs-fMRI and model both functional dependency among brain regions and the temporal dynamics of brain activity, spatio-temporal graph convolutional networks (ST-GCN) are applied to formulate functional connectivity networks in the format of spatio-temporal graphs \cite{gadgil2020spatio}.

\section{Bayesian Deep Learning \& Variational Inference}
\label{sec:bayes_dl}
Bayesian networks are statistical methodology that combines standard networks with Bayesian inference. Following the Bayes rule (Eq. \ref{eq:bayes_rule_vi}), the random variables of a problem can be represented as a directed acyclic graph known as \textbf{Bayesian Network} or \textbf{belief network} \cite{goodfellow_deep_2016}.
The Bayesian inference is an essential means of calculations across various disciplines, personalized advertising recommendation systems, in healthcare applications \cite{angelino2016}, the research in astronomy \cite{astronomy_2015}, and, search engines \cite{ms_bing_2010}.

Let \textbf{z} $= \{z_1,z_2,...,z_N\}$, and \textbf{x} $= \{x_1,x_2,...,x_M\}$ denote the latent variables and the observations respectively. The latent variables facilitate the representation of the observations' distribution. Given a prior distribution $p(\text{z})$ over the latent variables, the Bayesian model maps the latent variables to the observations by the likelihood function $p(\text{x|z})$. Thus producing the joint distribution of the latent variables and observations:
\begin{equation}
    \label{eq:bayes_rule_vi}
    p(\text{z,x}) = p(\text{x|z})p(\text{z}) = p(\text{z|x})p(\text{x})
\end{equation}
In Bayesian models, \textbf{inference} involves in calculating the \textbf{the posterior distribution} which is the conditional distribution of the latent variables given the observations:
\begin{equation}
    \label{eq:joint_zx_vi}
    p(\text{z|x}) = \frac{p(\text{z,x})}{p(\text{x})}
\end{equation}
The marginal density of the observations $p(\text{x})$ is called \textbf{evidence} which is calculated by integration over latent variables:
\begin{equation}
\label{eq:evidence_vi}
    p(\text{x}) = \int p(\text{x,z})d\text{z}
\end{equation}

\subsection{Hidden Markov Model}
\textbf{Hidden Markov Model} (HMM) is a probabilistic Bayesian network architecture \cite{Smyth1997ProbabilisticModels} that approximate the likelihood of distributions in a sequence of observations \cite{GHAHRAMANI2001ANNETWORKS}. As opposed to Bayesian networks, these networks are undirected and can be cyclic.
The family of HMMs, including the \textbf{Hidden Semi-Markov Model} (HSMM), are widely used to identify patterns in sequential data of time varying and non-time varying nature \cite{Selva2018EstimationModel}. They are well suited for sequencing time series problems with a linear degree of growth over data patterns \cite{Yu2010HiddenModels}. A generalized HMM is composed of a state model of Markov process $z_t$, linked to an observation model $P(x_t|z_t)$, which contains the observations $x_t$ of the state model. 
 
While HMMs are considered agnostic of the duration of the states, the HSMMs can take the duration of each state into consideration \cite{Murphy2002HiddenModels}, which makes HSMMs suitable for prognosis \cite{Dong2007AMethodology, 9659838}. Neither HMM nor HSMM can capture the inter-dependencies of observations in temporal data, which is a key factor in determining the state of the system. To overcome this shortcoming, one can use the \textbf{Auto-Regressive Hidden Markov Model} (ARHMM) which accounts for the inter-dependencies between consecutive observations to model longer time series \cite{ Dong2008AAR-HSMM,Guan2016EfficientModel,Mor2020AApplications}

HMMS can loose their efficiency when dealing with distributed state representations. The \textbf{Factorial HMM} (FHMM) is an extension of HMM that aims at addressing this problem by using several independent layers of state structure HMMs. These layers are free to evolve irrespective of the other layers, allowing  observations at any given time to be dependent on the value of all states at that time \cite{Ghahramani1997FactorialModels}. 

Due to exponential time complexity of this integration, its computation is intractable. Thereupon, the posterior distribution cannot be calculated directly. Rather, it is approximated \cite{blei_variational_2017}.
There are two major methods of posterior approximation:\par

\textbf{Sampling based}: Markov Chain Monte Carlo (MCMC) methods are often able to approximate the true and unbiased posterior through sampling, although they are slow and computationally demanding on large and complex datasets with high dimensions.\par
\textbf{Optimization based:} approaches for Variational Inference (VI) tend to converge much faster though they may provide over-simplified approximations.

The following subsections explain each of these approaches on these topics due to their importance.
\subsection{Markov Chain Monte Carlo}
\label{subsec:mcmc}
Monte Carlo estimation is a method for approximating the expectation of random variables where their expectation may involve intractable integrations as in Eq. \ref{eq:evidence_vi}. Markov Chain Monte Carlo, Metropolis-Hastings (MH) sampling, Gibbs sampling, and their parallel and scalable variations are instances of MCMC estimations \cite{angelino2016}.

Although the basic Monte Carlo algorithm requires the samples to be independent and identically distributed (i.i.d), obtaining such samples my be computationally intensive in practice. Nonetheless, the sample generation process can still be facilitated by satisfying some properties as described below \cite{Sharma_2017_mcmc}:

\textbf{Markov Property:} Given the past and present states, the probability of transition to the future states relies on the present state only. Mathematically speaking, a \textbf{Markov chain} is a sequence of random variables $\text{X}_1,\text{X}_2,...,\text{X}_n$ representing states, that hold the following property:
\begin{multline}
\label{eq:markov_mcmc}
    P(\text{X}_{n + 1}=x|\text{X}_n = x_n, \text{X}_{n-1} = x_{n-1} ,..., \text{X}_1 = x_1) = \\ P(\text{X}_{n + 1}=x|\text{X}_n = x_n)    
\end{multline}

\textbf{Time-homogeneity:} A stochastic process that the probability of transition is independent of the index $n$, is time-homogeneous.

\textbf{Stationary distribution:} A probability distribution of a Markov chain represented as a row vector $\pi$ that is invariant by matrix of transition probabilities K.
\begin{equation}
    \label{eq:stationary_mcmc}
    \pi = \pi \text{K}
\end{equation}

\textbf{Irreduciblity:} A Markov chain is irreducible if in a discrete state space, it can go from any state $x$ to any other state $y$ in a finite number of transitions. In mathematical terms, given that:
\begin{equation}
    \label{eq:irreducible_mcmc}
    \text{K(x,y)} = P(\text{X}_{n + 1}=y|\text{X}_n = x) 
\end{equation}
Where K is a matrix, there exist an integer $n$ such that $K_{(x,y)}^{n} > 0 $.

A stationary distribution of a chain is unique if it has stationary distribution and it is irreducible. Considering a Markov chain with a unique stationary distribution $\pi$, according to the law of large numbers \cite{Sharma_2017_mcmc},  the expectation value of a function $f(x)$ over $\pi$ can be approximated by calculating the mean of the outputs from the Markov chain:
\begin{multline}
\label{eq:mcmc_approx}
    E_{\pi}[f(x)] = \int f(x)\pi(x)dx = \underset{n \rightarrow \infty }{\text{lim}}\frac{1}{n}\sum_{i=1}^{n}f(x_i)
\end{multline}

For a more detailed explanation on MCMCs, the readers may consult \cite{Sharma_2017_mcmc,angelino2016}.

\subsection{Variational Inference (VI)}\label{subsubsec:variational}
Variational Inference (VI) is of high importance in modern machine learning architectures. Regularization through variational droput \cite{kingma_variational_2015,krueger_bayesian_2018}, representing model uncertainty in classification tasks and reinforcement learning \cite{gal_dropout_2016}, are a few of scenarios in which variational inference is utilized. The core idea in VI is to find an approximate distribution function which is simpler than the true posterior and its Kullback-Liebler divergence \cite{kullback1951} from the true posterior is as lowest as possible \cite{zhang_advances_2019}.

The problem changes to the search for a candidate density function $q_c(\text{x})$ among a specified family of distributions $D$ such that it best resembles the true posterior function:
\begin{equation}
    \label{eq:opt_vi}
    q_c(\text{z}) = \underset{q(z) \in D}{\text{argmin}} \text{ KL}(q(\text{z}) \| p(\text{z|x}))
\end{equation}
The Eq. \ref{eq:opt_vi} may be optimized indirectly through maximization of the variational objective function ELBO(q):
\begin{equation}\label{eq:elbo}
\text{ELBO(q)} = \mathop{\mathbb{E}}[\text{log }p(\text{x|z})] - \text{KL }(q(\text{z}) || p(\text{z}))
\end{equation}
where ELBO(q) is called \textit{evidence lower bound} function. The $\text{KL}(q(\text{z})||p(\text{z}))$ encourages the density function $q(\text{z})$ to get closer to the prior function. And the expected likelihood  $\mathop{\mathbb{E}}[\text{log }p(\text{x|z})]$ encourages preference of latent variable configurations that better explain the observed data.
The Eq. \ref{eq:elbo} may be rewritten as follows:
\begin{equation}
    \text{log }p(\text{x}) = \text{KL} (q(\text{z}) || p(\text{z|x})) + \text{ELBO(q)}
\end{equation}
The value of the left hand side (log-evidence) is constant and the $\text{KL(.)} \geq 0$. As a result, ELBO(q) is the lower-bound of evidence.

There are numerous extensions and proposed approaches for variational inference in the literature such as Expectation Propagation (EP) \cite{Seeger05_exp} and stochastic gradient optimization \cite{angelino2016}. For a more detailed and comprehensive review, the readers are encouraged to consult \cite{blei_variational_2017,zhang_advances_2019}.

\section{Convolutional Neural Network}
Convolutional Neural Networks (CNN) are the prevalent approach in extracting features from image data. Though several variants of CNNs have been proposed, they all share pretty much the same basic components: convolution, pooling, and fully-connected layers. \par
\textbf{Convolution Layer:} Extracts features from a given input layer and stores them on several feature maps which make up the higher layer. Each convolution layer has several feature extractors called kernels(filters) that each of them correspond to a single feature map. Every single neuron of the feature map corresponds to a group of neighboring neurons from the input layer referred to by neuron's receptive field. Each kernel is used to calculate the convolution over all of the possible receptive fields of the input layer. The convolution value is then passed through a non-linear activation function such as $tanh(.)$ or $sigmoid$ or $ReLU$ \cite{nair_rectified_2010} to add non-linearity to the representation. 
\par

The feature value $z_{i,j,k}^{l}$, at location (i,j) of the $k-$th feature map of layer $l$ can be calculated as:

\begin{equation}
\label{eq:conv}
    z_{i,j,k}^{l} = w_{l}^{k} \odot x_{i,j}^{l} + b_{k}^{l}
\end{equation}
Where $x_{i,j}^{l}$ represents the receptive field of neuron $z_{i,j,k}^{l}$ in the input layer. And the symbol $\odot$ represents the discrete convolution i.e. the sum of elements of Hadamard product(element-wise) of the two matrices. The activation of each feature can be obtained from the Eq. \ref{eq:conv_act} \cite{Million07thehadamard}.

\begin{equation}
\label{eq:conv_act}
    a_{i,j,k}^{l} = f(z_{i,j,k}^{l})
\end{equation}
Where $f$ refers to the activation function.
\par

\textbf{Pooling Layer:} The next step after convolution is reducing the size of the shared feature map. Various pooling operations are proposed; though the average pooling and max pooling are typically used \cite{gu_recent_2018}. The pooling operation can be represented mathematically as:
\begin{equation}
\label{eq:pool}
    y_{m,n,k}^{l} = pool(\{\forall a_{k}^{l} \in V_{m,n}\})
\end{equation}

The neuron $y_{m,n,k}^{l}$ at location (m,n) of the $k-$th pooled feature map of layer $l$ would be calculated from a set of neighboring neurons $V_{m,n}$ on the convolution feature map passed through the pooling function $pool$. One of the main advantages of the convolutions over other architectures are having the shift-invariance. A small displacement(rotation, translation) of the input wouldn't change the output dramatically. The main characteristic comes from sharing the kernels and pooling layers.

\par

\textbf{Fully-Connected Layer:} After several convolutional and pooling layers, that perform as feature extractors, typically a few fully-connected  layers (MLPs as discussed in section \ref{sec:anns}) are added in order to perform high-level reasoning given the extracted features \cite{zeiler_visualizing_2013}. For classification tasks the fully-connected network takes in all the neurons from the previous layers as input and provide an output of classes followed by a $softmax$ function. Given a dataset of $N$ pairs of inputs and outputs $\{(x_1,y_1),(x_2,y_2),...,(x_N,y_N)\}$, and the weights and biases of the whole network denoted by $\theta$, the total classification error of the network can be calculated by the following loss function: \par

\begin{equation}\label{eq:cnn_loss}
    \mathcal{L} = \frac{1}{N}\sum_{i=1}^{N}\ell(\theta;y_i,\hat{y}_i)
\end{equation}

Where $\hat{y}_i$ denotes the class label calculated by the network and $y_i$ is the true value of the class label. The training process of the network would involve the global minimization of loss function. The model's parameters $\theta$ can be updated using Stochastic Gradient Descent (SGD) \cite{wijnhoven_fast_2010} that is a common method for training CNNs, although various other optimization methods and loss functions have been proposed. Additionally, the fully-connected layers and the final layer of CNN may be replaced by some other types of networks or models. Also, numerous variants of CNNs and additional components are proposed in the literature \cite{zeiler_deconvolutional_2010,le_tiled_2010,lin_network_2014-1,yu_multi-scale_2015}. The readers can consult \cite{gu_recent_2018} for a comprehensive introduction to the CNNs.

CNN and its extensions can be seen in almost all of the state of the art methods of deep representation learning. It has demonstrated competitive capabilities in numerous supervised and unsupervised tasks on 2D/3D images and point cloud data \cite{liu_deep_2019} such as image retrieval \cite{cai_medical_2019}, segmentation \cite{long_fully_2015,ronneberger_u-net:_2015,chen_octopusnet:_2019}, registration \cite{miao_cnn_2016}, object detection \cite{zhao_object_2019,frustum_pointnet_2017}, and data augmentation \cite{yi_generative_2019,shin_medical_2018}. It has also been applied to  sequential data to extract longitudinal patterns of signals \cite{weng_representation_2019,parvaneh_cardiac_2019}, i.e. applicable to 1D data as well. CNNs, also empower reinforcement learning algorithms \cite{li_deep_reinforcement_learning_2018} and may be utilized to analyze graph data \cite{wu2019comprehensive,henaff2015deep} as will be discussed in section \ref{sec:gnn}.

\section{Word representation learning}
\label{sec:wordEMB}

Representing words numerically is one of the most crucial components of natural language processing. It can be viewed as the foundation of why we can employ ANN for NLP tasks. One-hot vector is the simplest way of representing a word in computer-readable format, assigning 1 for each word in a vector with a dimension of vocabulary size\cite{one_hot_define}. The main flaw in this approach is that it neglects the semantic relatedness between words. Vector space model \cite{vector_space}is one of the first methods of representing words mathematically that made it possible to calculate similarity between documents in the field of Information Retrieval.

More recent works use distributed hypothesis for numeric representations of words known as word embeddings. In modern NLP,\textbf{ word embeddings} have emerged as a method for learning low-dimensional vectors from text corpora in which the similarity between words is projected.
Word embeddings will not only present the semantic meanings of the word but also may show the word-context information. We can view language model as a tool which represents a sequence of words' probability distribution based on training data\cite{language_model}. 
A language model learns the joint probability function of a sequence of words in a corpus \cite{bengio2000neural}.
Word embeddings are dense, distributed, fixed-length word vectors \cite{word_emb_a_survey_2020,Word_embedding_understanding_natural_language:_A_survey} that are used to estimate probability distributions of words in corpora for prediction-based neural language models. \cite{bengio2000neural} which can be referred to as one of the earliest Language models on the neural net, is the basis for most of the current works in NLP. \cite{bengio2000neural} found it hard to learn the joint probability function of sequences of words because of the curse of dimensionality. They proposed a method that enable the model to learn about an exponential number of semantically neighboring sentences by estimating the distributed representation for each word. In this model, the learned distributed encoding of each word is fed into the last unit(softmax) to predict what the probabilities are for incoming words. There are several other types of research done for word embeddings in prediction models \cite{Mnih2007ThreeNG}\cite{A_unified_architecture_natural_language_processing}\cite{From_word_to_sense_embeddings:_A_survey}.

Later on, as one of the most popular approaches, Word2vec \cite{mikolov2013} has proposed two methods: CBOW (continuous bag of words) and skip-gram(SG)\cite{mikolov2013linguistic,pennington-etal-2014-glove}.
In "Continuous Bag of Word," the model predicts the middle word given the distributed representations of context (or surrounding words). In contrast, SG predicts the context words given the center word. Afterward, they proposed an improvement to circumvent the intractable computational burden of both methods by using negative sampling \cite{Mikolove_negative_sampling}. In negative sampling method, for calculating the softmax equation denominator, they use a random subsample of frequent words instead of the entire training set.  

The cross-entropy loss function is $ H (p,q)= - \sum_{x \in X} p(x)log^ {q(x)}$ in this model

\begin{equation}
P(\text { out } \mid \text { center })=\frac{\exp \left(u_{\text {out }}^T v_{\text {center }}\right)}{\sum_{w \in V} \exp \left(u_w^T v_{\text {center }}\right)}
\label{eq:skip-gram}
\end{equation}

 As another influential approach Glove \cite{pennington-etal-2014-glove} captures the difference between a pair of words as ratio of co-occurence probabilities for target words with selected context word.
 Later on, FastText \cite{fast_text} built upon the "Glove" and "Word2Vec" to mitigate their shortcoming in handling out-of-vocabulary(OOV) \cite{KHATTAK2019100057}\cite{athiwaratkun2018probabilistic}.

\section{Sequential Representation Learning}
\label{sec:seq_rl}
In many applications, the data is structured as sequences with different lengths in which the order of the elements in sequences is important. For instance, the sentences in natural language processing (NLP) \cite{fawaz_2019_time_series}, or the records of medical data \cite{ching_opportunities_2018,Johnson2016}. Capturing and representing the patterns of the features along time requires an architecture that is able to: a) process inputs of different lengths. b) capture the possible dependency of the data points on one another. Recurrent Neural Networks (RNN) \cite{rumelhart_learning_1986}, provide both of these capabilities by sharing parameters to the next step\cite{goodfellow_deep_2016}. The following sections discuss the major variants of RNNs.

\subsection{Recurrent Neural Network}
The general architecture of a recurrent neural network is composed of cells with hidden states. In mathematical terms, hidden units $h_t$ store the state of the model that depends on the state at previous time step $h_{t-1}$ and the input of the current time step $x_t$:
\begin{equation}
\label{eq:rnn_hidden}
    h_{(t)} = f_a(Wh_{(t-1)},Ux_{(t)} + b)
\end{equation}
Where the matrices $U,W$ are weight matrices and $b$ is bias vector, and $f_a$ represents the activation function. The same set of model parameters is used for calculation of $h_t$ for any of the elements in a sequence of inputs $(x_1,x_2,...,x_n)$. In this way, the parameters are shared across the input elements.  For supervised tasks such as classification, the hidden unit $h_t$ is mapped to the output variables $y_t$ via the weight matrix $V$:
\begin{equation}
    \label{eq:rnn_out}
    \hat{y_t} = \text{softmax}(Vh_t + c)
\end{equation}
$c$ is the bias vector. Due to the recursive nature of Eq. (\ref{eq:rnn_hidden}), the unfolded computational graph for a given input sequence, can be displayed as a regular neural network.

RNNs can be trained by back-propagation \cite{goodfellow_deep_2016} as if it is calculated for the unfolded computational graph. Two of the most important problems with RNNs are vanishing and exploding gradients. The longer the input sequence, the more gradient values are multiplied together, which may cause it to converge to zero, or exponentially gets large. Either way, the RNN fails to learn anything. Various methods have been proposed to facilitate training RNNs on longer sequences. For instance, skip connections \cite{skip_connections_lin1996} let the information flow from a farther past to the present state. Another method, is incorporation of leaky units \cite{leaky_units_bengio1996} in order to keep track of the older states of hidden layers by linear self-connections. Nonetheless, the problem of learning long-term dependencies is yet to be resolved completely.



\subsection{Long Short-Term Memory (LSTM)}
The most prominent architecture for learning from sequential data, is Long Short-Term Memory that combines various strategies for handling longer dependencies \cite{lstm_1997}. In LSTM, tuning of the hyper-parameters is a part of the learning procedure. The general architecture of an LSTM cell contains separate gates that control the information flow across the time steps of sequences:

\textbf{Input Gate:} Controls whether the input is accumulated into the hidden state.

\textbf{Forget Gate:} Controls the amount of effect that the previous state has on the current state. Whenever this gate lets information flow in completely, it acts as a skip-connection \cite{skip_connections_lin1996}. Otherwise, similar to leaky units, it keeps track of previous hidden states with a linear coefficient.

\textbf{Output Gate:} The output of the LSTM can also be stopped. 
\subsection{Gated Recurrent Unit (GRU)} 
Another variant of recurrent neural network that learns long term dependency through gates is Gate Recurrent Unit \cite{GRU_cho_et_al2014}. The structure of GRU is simpler compared to LSTM. Hence, there are fewer parameters in a GRU cell. There are only two types of gates in a GRU cell:

\textbf{Update Gate:} Controls the weights of interpolation of the current state and the candidate state in order to update the hidden state.

\textbf{Reset Gate:} Makes the hidden state forget the past dependencies.

All the gates in variants of the gated recurrent neural networks, are controlled by linear neural networks. Training the recurrent neural network as a whole, also trains and updates the weights of the gate controller networks.

\subsection{Applications}
There are myriad problems and use cases for the RNNs. Instances are: analysis and embedding of texts and medical reports to be combined with medical images \cite{hsu_unsupervised_2018},  real-time denoising of medical video \cite{sadda_real-time_2018}, classification of electroenephalogram (EEG) data \cite{craik_deep_2019}, generating captions for images \cite{vinyals_show_2015,hossain_comprehensive_2019,shin_learning_2016,soheyla_image_caption_review_2019ASR}, biomedical image segmentation \cite{chen_combining_2016}, semantic segmentation of unstructured 3D point clouds  \cite{xiaoqing_2018_context_3d_cloud,huange_slice_networks_3d_seg}. Other examples of RNN applications are predictive maintenance \cite{baghaei_prediction_prognostics_2020}, prediction and classification of ICU outcomes \cite{retainvis_2018,Barbieri2020_rnn, baghaei_sepsis2019}.

\section{Transformers}
Transformer\cite{vaswani2017attention}, is a sequence to sequence model that was a turning point in natural language processing and then became a dominant architecture in other fields like computer vision\cite{touvron2021training,dosovitskiy2020image}, speech recognition\cite{Dong2018SpeechTransformerAN,karita19_interspeech,T_end_to_end}, video understanding \cite{ramesh2021zero}, long medical prescription information extraction \cite{beltagy2020longformer}, and several other use cases\cite{ramesh2021zero,tan2019lxmert,zhou-etal-2018-multi}. Attention-based encoder-decoder models are innovated to solve the shortcomings of RNN, LSTM, and GRU, which were fairly known as the state-of-the-art approacehes. Transformers mainly learn the dependency between sequence elements in a long-term context. \leavevmode\\

\subsection{Learning to align and translate}\leavevmode

\begin{figure}[htbp]
\centering
\includegraphics[width=1\linewidth]{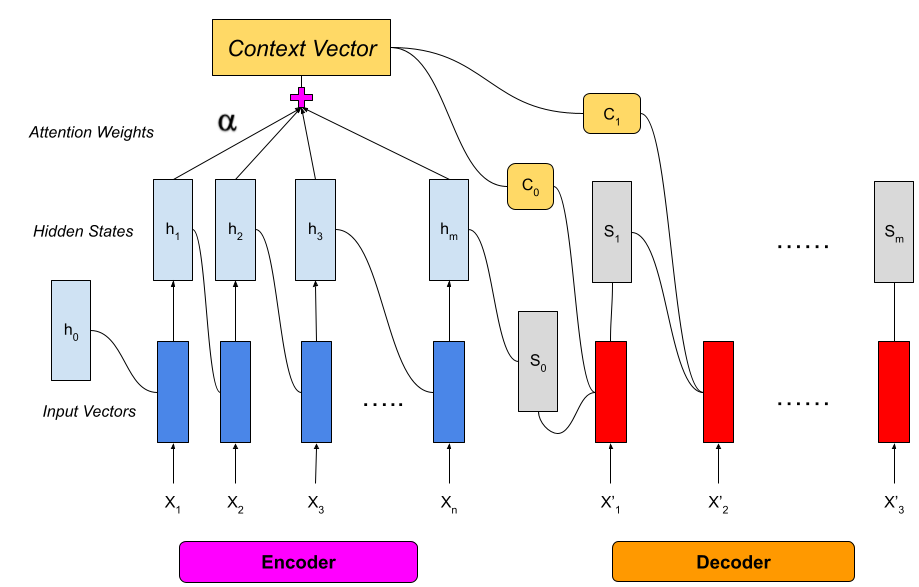}
   
\label{fig:attmodel}
\caption{Based on the current target state $h_{t}$ and all source states $h_{s}$, the model determines an alignment weight vector at time step $t$. A global context vector $c_{t}$ is then computed as the weighted average over all the source states\cite{bahdanau2014neural}}
\end{figure}

The first Encode-decoder model with an attention mechanism was proposed by \cite{bahdanau2014neural} in 2015 as a novel architecture to improve the performance of neural machine translation models. The main contribution of Badanau et al.\cite{bahdanau2014neural} was an extension(called attention) on the decoder part  by calculating the weighted sum of the inputs' hidden states. In basic encoder-decoder, we have a single encoded fixed-length vector, while, Badanau et al encode variable-length vectors and, in the decoding, part performs selectively.\\
As it is depicted in the figure \ref{fig:attmodel}, the $c_{t}$ is what is added to this model as attention. So, the $ s_{t}=f(s_{t-1}, y_{t-1}, c_{t}) $ output of the decoder will be based on the $ c_{t}= \sum\alpha_{tt'}h_{t}$  which is defined as weighted attention where 

\begin{equation}
\alpha_{t t^{\prime}}=\frac{\exp \left(e_{t t^{\prime}}\right)}{\sum_T \exp \left(e_{t T}\right)}
\end{equation}

and $e_{tt'}$ is called alignment score. In other words, $ \alpha_{tt'}$ is called amount of attention $y^t$ (the output in time step t) pay to $x^{t'}$. There are different options to calculate the alignment score, and it is one of the parameters can be trained, but in general, it is based on $align(h_{i},s_{0})$. \\

This model was a novel approach and inspired recent state-of-the-art models; however, It was bound to the limitation of traditional encoder-decoder recurrent models and did not have parallel computing.

\leavevmode \\
\subsection{Attention is all you need}
\leavevmode \\
Despite the great invention by Badanau et al.\cite{bahdanau2014neural}, which mainly was an add-on for RNN-based models, this model was still challenging to train because of the long gradient path, specifically for long data sequences. In 2017 google introduced transformers\cite{vaswani2017attention} that takes advantage of attention on the relative part of the input sequence. This model solves the two main drawbacks of LSTM-based models: 1. parallel computing, 2. Long gradient path. This model first applied to natural language processing tasks, became pervasive, and significantly impacted many other fields. 

The model architecture is based on multiple same-structured encoder-decoders stacked on top of each other. As stated earlier, this architecture was proposed for NLP tasks. The first step is passing the input through the \textit {'input embedding'} layer to transform one-hot token representation into a word vector; after positional encoding, feed the result to the encoder. The central part of the encoder and decoder blocks is characterized by a multi-headed self-attention mechanism  $(Q, K, V)$ which the result passes through point-wise feed-forward networks.\\ 

\textbf{Self-attention structure:}\leavevmode \\

Given a series of inputs, learn self-alignment to estimate the relevance of one element to all others in the series. \\
\textbf{Step 1:} Randomly generate $W_{Q}$, $W_{K}$, $W_{V}$ weights and calculate:
\leavevmode \\ $Queries=X_{k}$ $ \scriptstyle(\text{embedding}) $ $*W_{Q}$ \leavevmode \\
$ Key =  X_{k}$ $ \scriptstyle(\text{embedding}) $ $*W_{K}$  \leavevmode \\
$ Value =  X_{k}$ $ \scriptstyle(\text{embedding})  $ $*W_{V}$ \leavevmode \\

$ \scriptstyle(\text{Where  $X \in \mathbb{R}^{T\times d_{m}}$, $W^{Q} \in \mathbb{R}^{D\times D_{Q}}$,$W^{K}\in \mathbb{R}^{D\times D_{K}}$, $W^{V}\in \mathbb{R}^{D\times D_{V}}$}) $

\textbf{Step 2:}\leavevmode \\
Calculate z score for each input by applying row-wise softmax on the score between each input
\begin{equation}
Z(Q,K,V)= \textit{Softmax}( \frac{Q.K^T}{\sqrt{d_{K}}}).V
\end{equation}
Concatenate the $Z$ scores, initialize a new weight matrix and multiply it by the $Z$ score. Finally, feed the result to the FCNN.
\\
Keys and queries can be considered as the information that each token wants to share and gather with and from others.

\subsection{Extra Large Transformers:}
Transformers have become indispensable to the modern deep learning stack by significantly impacting several fields. This has made it the center of focus and caused a dizzying number of model variants proposing basic enhancements to mitigate a widely known concern with self-attention: the quadratic time and memory complexity \cite{survey-transformer}. These two drawbacks can be a significant obstacle to model scalability in many settings. Approaches to mitigating this limitation can be classified in many directions\cite{survey-of-transformers}, some of which are: 
\\
\par
\textbf{Recurrence:}
Transformer-XL\cite{dai-etal-2019-transformer} is one among the foremost famed extensions to vanilla Transformer and is a segment-level recurrence method connecting multiple adjacent blocks with a recurrent mechanism. This model introduces two key ideas: First, a segment-level recurrence makes it possible to cache and reuse hidden states from the last batch. The second may be a novel positional encoding scheme that allows temporal scheme coherence. 
As an extension to the block-wise approach, Transformer-XL splits the input into small non-overlapping subsequences referred to as blocks\cite{Fournier2021APS}. Even though it has impressive performance compared to the vanilla transformer, this model does not maintain long-term dependency and discards past activations as it runs through blocks. Mainly, Transformer-XL propagates the gradient across the current segment and caches it, then processes the second segment using the first segment memory(No-grad for the first segment), then goes to the third segment while discarding the first windows' gradient. Henceforth, we can view it as truncated back-propagation through time (BPTT).
The main point distinguishing this model from others is the relative positional encoding scheme that allows temporal coherence. The relative positional encoding encodes distances on edges rather than nodes. While there was previous work on relative positional encoding\cite{shaw2018self}, two features is added: first global content and location bias. Second, sinusoid embedding replaces the trainable one. Their result shows that Transformer-XL performs better than vanilla transformers even without a recurrence mechanism. Compressive Transformer \cite{pmlr-v80-parmar18a} is another model which can be classified as a recurrence approach. 
\\

\par 
\textbf{Reduced Dimenssions/ Kernels/ Low- rank methods:}
Transformers are RNNs\cite{transformers-are-RNNS} that use a kernel with the function $\phi(X)=elu(x_{i})+1$ instead of softmax that maps the attention matrix to its approximation. The function is applied on 'Keys' and 'Queries,' lowering their dimension. ${Q}_{({\tiny N}\times D)}.{k^T}_{(D\times N)}$ and avoid computing the ${N\times N}$ matrix. Linformer\cite{wang2020linformer} and Synthesizer\cite{tay2021synthesizer} are other models based on this approach. 

\leavevmode \\
\par 
\textbf{Sparse attention}
Longformer\cite{longformer-cite} with linear complexity of O(n) leverages a global memory technique and is analogous to applying CNN. A combination of sliding window and global attention technique is applied to each Query to reduce dimensions. 
Longformer, Bigbird\cite{big-bird}, ETC\cite{ainslie-etal-2020-etc}, and SWIN Transformer\cite{9710580} are all in the same category using sparse attention techniques. Image Transformer \cite{pmlr-v80-parmar18a} and Axial Transformer\cite{ho2020axial} are other extended sparse attention works focused on vision data

\subsection{Pre-trained Models}
\leavevmode \\
Pre-trained models are neural networks previously trained on large-scale corpus representations designed to be capable of transfer and fine-tune on different downstream NLP tasks.
Word embeddings as a base that enabled us to utilize machine learning for processing natural language can be viewed as pioneers of widely used pre-rained representations. Word2vec\cite{mikolov2013}, and Glove \cite{pennington-etal-2014-glove}, which we discussed earlier, are among the most famous models learning a constant embedding for each word in vector space. In what follows, we will try to pinpoint the most famous and important pre-trained models that retain contextual representations, which those mentioned earlier are incapable of.
 
\cite{NIPS2015_7137debd} from google in 2015 is one of the earliest supervised sequence learning based on LSTM that pre-trained the whole language model to be used on different classification tasks.
ELMO\cite{peters-etal-2018-deep} deep contextualized word representation is analogous to the earlier one; however, it was bi-directional. CoVE\cite{cove} is another recurrent based in this category which has shown good performance. GPT1 \cite{gpt1} by OpenAI is a Transformer-based pre-trained model trained on a large book corpus dataset to learn a universal representation that can transfer with little adaptation. "Deep Bidirectional Transformers for Language Understanding " - BERT \cite{bert}, a well-known turning point in the NLP area (perhaps also the entire ML stack), trains left context and right context at the same time rather than doing individually and concatenating at the end. As BERT has access to both directions, they mask out K\% of the sequence to prevent cheating. XLNet\cite{xlnet} RoBERTa\cite{liu2020roberta},ERNIE \cite{Ernie}, ELECTRA \cite{Clark2020ELECTRA},  are numbers of BERT variation. 

All the mentioned models have proven effective in natural language processing studies. These successful observations within the NLP space inspired researchers to apply a similar approach to other domains. \cite{wang2020linformer} has shown that pre-trained models' success is not limited to transformer-based ones. They have demonstrated their pre-trained convolution seq2seq model can beat pre-trained Transformers in machine translation, language modeling, and abstractive summarization. Vision Transformer(ViT) \cite{dosovitskiy2021an} is one of all the foremost recent pre-trained models by google that helps transfer learning in image classification tasks. This model has shown outstanding results in training a pure transformer applied directly to sequences of image patches. ResNet50 \cite{resnet50}, a pretrained CNN-based model which allows training networks with up to 1000 layers. ResNet50 consists of a succession of convolutional layers with different kernel settings. \cite {liu-paper} \cite{1404.5997} \cite{Szegedy_2016_CVPR}\cite{Sandler_2018_CVPR} are all trained on the vast number of datasets for various image classification transfer learning usage categories. 

While these models have shown excellent results, the range of pre-trained models is not restricted to the mentioned. One is to precisely study and examine different models and approaches to search out their dataset's best and most efficient model. 

\subsection{recurrent cell to rescue:}
After GPUs became accessible as a computation power in the ML toolkit, LSTM emerged as a practical approach in various sequence-based machine learning models. With the introduction of word embeddings in 2013, LSTM and other RNN-based models have been widely dominant in sequence learning NLP problems. After presenting transformers with their \textit{All-to-all comparison} mechanism and their performance on transfer learning tasks, they became the SOTA model and dominated the NLP and, subsequently, deep learning space. 

While transformers can grasp the context and be used more efficiently for transferring knowledge to tasks with limited supervision by pre-trained models, these benefits come with quadratic memory and time complexity of $O(N^2)$ \cite{transformers-are-RNNS}. Most of the current pre-trained transformer-based models do only accept 512 numbers of the input sequence. IndRNN model \cite{IndRNN} has shown the ability to process sequences over 5000 time steps. The \textit{Legendre Memory Unit} \cite{legendmemory} is based on recurrent architecture and can be implemented by a spiking neural network\cite{accurate-neo}, which can maintain the dependencies across 100,000 time steps. Apart from computation cost, \cite{wang-etal-2020-pretrain} by Facebook shows that the accuracy gap between Bert-based \cite{bert} pre-trained models versus vanilla LSTM for a massive corpus of data is less than 1\%. Henceforth, a competitive accuracy result is achievable by training a simple LSTM when many training examples are available. They also show that reusing the pre-trained token embeddings learned in BERT can significantly improve the LSTM model's accuracy. \cite{schneider2021re} shows that standard transformers are not as efficient as RNN-based models for reinforcement learning tasks. [A-13] has investigated the performance of Transformer and RNN in speech application and shows both have the same performance in text-to-speech tasks and slightly better performance by Transformer in the automatic speech recognition task.

On the other hand, \cite{schneider2021re} shows that their attention-based model can outperform the state-of-the-art in terms of precision, time, and memory requirements for satellite image time series. \cite{Pretrained-Transformers-As-Universal} has compared LSTM performance with transformers in their proposed Frozen Pretrained Transformer model as part of their paper. They evaluate a diverse set of classification tasks to investigate the ability to learn representations for predictive learning across various modalities and show that transformers perform better. \cite{SeTransformer} has proposed an improved-Transformer-based comment generation method that extracts both the text and structure information from the program code. They show that their model outperforms the regular Transformer and classical recurrent models. \cite{Transformer-Based-Direct-Speech-To-Speech} is a transformer-based transcoder network for end-to-end speech-to-speech translation that surpasses all the SOTA models in natural speech-to-speech translation tasks.\cite{li2021act} has introduced an AttentiveConvolutional Transformer which takes advantage of Transformer and CNN for text classification tasks. Their experiment reveals that ACT can outperform RNN-based models evaluated on three different datasets.

\textbf{Combining Recurrent and Attention:}
R-Transformer\cite{wang2020rtransformer} inherits the Transformers' architecture and is adding what they call "Local RNN" to capture sequential information in data. The main improvement proposed is defining a sequence window to capture the sequential information and sliding the Local RNN over the whole time series to get the global sequential information\cite{FELLNER2022100851}. This approach is similar to 1-D CNN; however, CNN ignores the sequential information of positions. Also, the Transformer's positional embedding that mitigates this problem is limited to a specific sequence length. Henceforth, they have proposed a 'Local RNN' model that can efficiently do parallel computation of several short sequences to capture the local structure's global long-term dependency by applying a multi-head attention mechanism. This model has replaced the Transformers' position embeddings with multiple local RNNs, which can outperform the simple recurrent approaches such as GRU, LSTM, convolutional \cite{Empirical-Evaluation-of-Generic-Convolutional}, and regular Transformer. \cite{SeTransformer} has proposed a modified LSTM cell to mitigate the similarity between hidden representations learned by LSTM across different time steps in which attention weights cannot carry much meaning. They propose two approaches: first, by orthogonalizing the hidden state at time $t$ with the mean of previous states, they ensure low conicity between hidden states. The second is a loss function in which a joint probability for the ground truth class and input sentences is used and also minimizes the conicity between the hidden states. These mutations provide a more precise ranking of hidden states, are better indicative of words important for the model's predictions, and correlate better with gradient-based attribution methods. 

\leavevmode \\
While we have mentioned works in which LSTM outperforms Transformers and vice versa, one should study the proper approach based on the dataset, accessible computation resources, and so forth.

\section{Transfer Learning}
\label{sec:trans_learning}
Many of the advancements in machine learning techniques make a huge improvement over the existing benchmarks. There are, however, some assumptions and challenges that make it difficult to apply the methods to real-world situations. In many cases, the assumption is that the trained model will be tested on the same feature distribution as the training stage. This assumption usually does not hold as the environment changes. In addition, many promising results are obtained by training models with large datasets. These pre-requisites makes it very challenging to adapt to many different tasks. For many applications, acquiring large amounts of data can be costly, time-consuming or even impossible. The absence of data for specific tasks may not be the only challenge; Massive data collection poses a huge privacy problem in many healthcare and medical applications \cite{ijerph18010271}. In other cases, annotating the data would require an expert and could be expensive, such as low-resource languages\cite{10.1145/3491102.3517639}. 

Transfer learning aims to alleviate the mentioned problems. Generally, transfer learning refers to when a learner wants to improve the performance on the target domain by transferring knowledge from the source domain. It derives from the human intuitive ability to share knowledge across different domains and tasks. For example, learning a language might help you learn the second one if there's some relation in between. The term itself is very general and there have been many extensions to it in recent years.

Transfer learning enables machine learning models to be retrained and reuse their previously learned knowledge. A general definition of the problem is divided into two components: \textit{Domain} and \textit{Task} \cite{pan_survey_2010}. 

The \textbf{Domain} is defined as $D =\{\chi,P(X)\}$, with $\chi$ representing the feature space, and $P(X)$ for each $X = \{x_1, ..., x_n\} \in \chi$ denoting the marginal probability over the feature space. In cases where different domains are encountered, the  source domain $D_S$ and the target domain $D_T$ can assume different feature space or marginal probability distributions \cite{pan_survey_2010}.

Given a specific domain $D =\{\chi,P(X)\}$, \textbf{Task} is represented by $T=\{y, f(x)\}$, where $y$ is the feature space and $f(x)=P(Y|X)$ denotes the function that can be learned from the training data to predict the target, in a supervised manner from the labeled data ${x_i , y_i }$, where $x_i \in X$ and $y_i \in Y$. In cases where no labels are considered for the data, as is the case for unsupervised algorithms, $y$ can be a latent variable such as the cluster number, or a variable that is produced by an unsupervised algorithm (e.g., the reduced dimensions of the original data)\cite{pan_survey_2010}. In light of both the domain D and the task T being defined as tuples, four transfer learning scenarios can be arise.

The first scenario is when the source and target domain are different ${X_s \neq Y_t}$. A good example of this is, in the computer vision community, where the source task is an image of a humans, but the target task is an image of an objects. A similar example can be found in NLP when it comes to cross-lingual adaptation.

The second scenario happens when $P(X_s) \neq P(X_t)$, the marginal probability distributions of source and target domain are different. This scenario is generally known as domain adaptation. An example could a detection problem where the source and target has different kind of cars.

The third occurs when $Y_s \neq Y_t$, the label spaces between the two tasks are different. For example, consider a detection problem where the source task considers the detection of cars, while the target task considers animals.

The last is when $P(Y_s|X_s) \neq P(Y_t|X_t)$, the conditional probability distributions of the source and target tasks are different. The imbalancness of data between source and target tasks is a very common example.

A number of surveys have been conducted to categorize the available methods\cite{5288526,https://doi.org/10.48550/arxiv.1911.02685,pan_survey_2010}. In \cite{5288526} the available methods are categorized into three different sections, transductive, inductive, and unsupervised transfer learning. In \cite{https://doi.org/10.48550/arxiv.1911.02685} the available methods are categorized in more detail based on the data or model perspectives. Although this categorization could give some insights, there are many newer methods that cannot fit in those categories or belong to more than one, zero-shot transfer learning \cite{chang2008importance}, reinforcement transfer learning \cite{https://doi.org/10.48550/arxiv.2009.07888}, and online transfer learning \cite{ZHAO201476} are among these methods.

Several sub-categories of transfer learning can be considered for each of these main categories based on the nature of the knowledge transfer. In the following subsection, the most prominent sub-categories of transfer learning will be discussed.


\textbf{Instance-based:} Despite differences in the source and the target domains, an instance based transfer learning, such as TrAdaBoost \cite{Dai2007Boosting} or Bi-weighting Domain Adaptation (BIW) \cite{wan2011bi}, adjusts the weights used for a subset of the source instances that are similar to the target domain, to predict the target instances. Since similarity of the selected source instance to those of the target domain play a crucial role in instance based transfer learning, a filter is used to remove dissimilar instances that would otherwise mislead the algorithm \cite{jiang2007instance, liao2005logistic, wu2004improving, pan_survey_2010, weiss_survey_2016, tan_survey_2018}.

\textbf{Feature-based:} Determining the common denominator between related tasks would allow for defining a representative feature that would apply to all domains and reduce differences between them. In this case, the common feature attempts to identify some partial overlap between the defined tasks. Having a representative feature among different tasks would also allow for a reduction in the overall error \cite{8460989,pan_survey_2010,REN2019106906,8372248}. While the source and target domains may have differences between them in their original data space, it is likely that the two would exhibit similarities in a transformed data space. Mapping-based deep transfer learning techniques, such as Transfer Component Analysis \cite{pan2010domain}, create a union between the source and target domain instances by applying a mapping between the two and transforming them into a new data space based on their similarity so that they can be used for deep nets \cite{tan_survey_2018,https://doi.org/10.48550/arxiv.2109.09214}.

\textbf{Network-Based:}
Different models derived from related tasks can have many similarities and differences. Similar models often have knowledge about the model parameters or the behavior of hyperparameters shared between the individual models. In such cases, it is possible to create a learning algorithm that infers the model parameters and the distributions of its hyperparameters by examining the prior distributions of several other tasks \cite{pan_survey_2010,lockner2021induced,YANG2020103199}. Similar to the learning and inference process followed by the human brain, where the trained brain cells can ad-hoc to other brain cells in related tasks, the network-based approach aims at using an already trained neural network as part of a much more extensive deep neural network. This approach trains the subnet on its relevant domain data, and the resulting pre-trained network is transferred to a larger deep net \cite{lockner2021induced,8360102,8008420}. A few examples of network-based deep transfer learning approaches include  ResNet, VGG, Inception, and  LeNet, which can extract a versatile set of features in the network's front layers \cite{tan_survey_2018}.

\textbf{Relational Knowledge-Based:} 
There are several instances, such as the social network data, where the data are not independent and identically distributed (IID). Relational domains allow for the handling of this scenario. In a relational domain, each entry is represented by multiple relations, not just a single identifier\cite{Xue_2021_ICCV}. Unlike other methods discussed before, the cross-domain relational knowledge transfer algorithms, such as TAMAR, use the Markov Logic Networks (MLNs) to transfer the relational knowledge without requiring each data point to be IID \cite{pan_survey_2010,7373394,rs11111358}. 

\textbf{Adversarial-Based:} 
Built on the strong foundation of the Generative Adversarial Networks (GANs), the adversarial-based approaches to transfer learning use a generator challenged by a discriminator to identify the transferable representations. A representation is considered transferable when it discriminates between the different components of the main learning task but does not discriminate the source domain from the target domain \cite{tan_survey_2018}. Most approaches use a single domain discriminator to align the source and target distributions, or use multiple discriminators to align subdomains \cite{https://doi.org/10.48550/arxiv.1711.02536,https://doi.org/10.48550/arxiv.1809.02176,https://doi.org/10.48550/arxiv.1702.05464,8842881}.

There is no unified approach where one can use Transfer learning. A very common transfer learning approach is when your target domain does not have sufficient training data. The model first pre-trains on the source data and then fine-tunes on the target data. Many well-known architectures in different communities are being used for related downstream tasks. In NLP, Bert \cite{devlin-etal-2019-bert}, Word2vec \cite{https://doi.org/10.48550/arxiv.1301.3781}, and ERNIE \cite{https://doi.org/10.48550/arxiv.1905.07129} are the famous models where a lot of downstream tasks can learn their specific related task with the shared knowledge backbone. Similarly, in the Vision community, Resnet \cite{https://doi.org/10.48550/arxiv.1512.03385}, Vision Transformers \cite{https://doi.org/10.48550/arxiv.2010.11929} and ConvNeXt \cite{liu2022convnet} could be used. Also in Speech, Wav2Vec \cite{https://doi.org/10.48550/arxiv.1904.05862}, DeepSpeech \cite{https://doi.org/10.48550/arxiv.1412.5567} and HuBERT \cite{https://doi.org/10.48550/arxiv.2106.07447} are among famous models. It is important to note that there are different levels of fine-tuning. With enough data in the target domain, fine-tuning can also alter the entire backbone representation. In many applications, however, this can be done partially (the last few layers) or just for the task-specific heads without changing the backbone representation. The mentioned models these are expected to be general enough so that they can be used for many downstream processes. For instance, Resnet trained to classify images into 1000 different categories. If the model knows to classify cars, the knowledge can be used to detect airplanes or even a different task like semantic segmentation\cite{https://doi.org/10.48550/arxiv.1702.08502}.

Other common ways of transfer are when the task is the same, but the domain changes. A helpful example might be applying the knowledge gained from the simulation data to real-world data \cite{https://doi.org/10.48550/arxiv.2009.13303}. In many applications like robotics, and computer vision, acquiring simulation data is very easy and straightforward.

The goal of transfer learning is to adapt the knowledge learned in one domain to another but closely related one. Many recent papers \cite{raghu2019transfusion,https://doi.org/10.48550/arxiv.2006.06882} suggested that using pre-trained models and fine-tuning might not be the optimal approach. It is therefore important to know why and what to transfer. Another issue with pre-training solutions is the accumulation of parameters in each sub-task. These networks can have millions or even billions of parameters \cite{https://doi.org/10.48550/arxiv.2006.06882}, it is then very impractical to fine-tune the backbone representation for every downstream task. Consider an application where the model should use the representation to do sentiment analysis along with entity recognition. If one wants to fine-tune a separate backbone for every downstream task, it would be very memory inefficient. Multitask Learning\cite{https://doi.org/10.48550/arxiv.1707.08114} aims to learn a shared representation for multiple related tasks which can be generalized across all tasks. As opposed to creating an instance of the backbone for each task, the representation is being shared across multiple tasks to improve efficiency \cite{https://doi.org/10.48550/arxiv.2110.04366}.

There are also instances where transfer learning occurs in the feature space; instead of transferring the representation to the new task, a related but fixed representation can be used. In this case, the main representation remains intact and a small network learns the representation specific to the target task. Having common latent features acts as a bridge for knowledge transfer. In \cite{nejatishahidin2022object} the authors trained a lightweight CNN module on top of a generic representation called mid-level representation. In comparison to training a complex CNN module which also learns the representations, they achieved superior performance in terms of accuracy, efficiency, and generalization with the method.

\subsection{Applications}
As mentioned, the use of transfer learning does not follow any conventional approach. Therefore, one should precisely study examples of how researchers can use transfer learning in their problems.  

When it comes to medical applications, both privacy and expert labeling are key issues that make data availability difficult. In \cite{alzubaidi2020towards,chen2019med3d,s19112645,https://doi.org/10.48550/arxiv.1912.12452} the authors try to transfer the knowledge learned from the pre-trained models, Resnet\cite{resnethe2016deep} or ALexNet \cite{NIPS2012_c399862d} trained on ImageNet, and transfer it for different tasks like Brain Tumor Segmentation, 3d medical image analysis and Alzheimer. \cite{cancers13071590} found that due to the mismatch in learned features between the natural image, e.g., ImageNet, and medical images the transferring is ineffective and they propose an in-domain transferring approach to alleviate the issue.

There was a great deal of success with transfer learning in the field of natural language processing. Several reasons exist for this, but largely it is because it is easy to access the large corpus of texts. It is inherent for pre-trained models to generalize across many domains due to the millions or billions of text data that they are trained on \cite{https://doi.org/10.48550/arxiv.1906.08237,bert,https://doi.org/10.48550/arxiv.2005.14165,https://doi.org/10.48550/arxiv.1910.01108,https://doi.org/10.48550/arxiv.2108.05542}. These representations can be transferred in different areas such as sentiment analysis \cite{zhang-etal-2021-cross,nayel-etal-2021-machine-learning,zhou-etal-2020-sentix}, Question Answering \cite{https://doi.org/10.48550/arxiv.1905.00537,guo-etal-2021-multireqa,yu-etal-2020-technical}, and Cross-lingual knowledge transfer\cite{nguyen-etal-2021-crosslingual,zhang-etal-2021-cross,do-gaspers-2019-cross}.

There are many speech recognition applications that are similar to NLP because of the nature of language. These applications are discussed in \cite{https://doi.org/10.48550/arxiv.2103.05834,9428334,SERTOLLI2021101204,8639655}

The progress of transfer learning in various domains have motivated researchers to adapt explored approaches for time series datasets\cite{9646532}. For time-series tasks, transfer learning applications range from classification\cite{https://doi.org/10.48550/arxiv.2207.07897}, anomaly detection \cite{XU2021107015,9328227} to forecasting \cite{YE2021107617,10.1007/s10489-020-01871-5}.

Transfer learning has also been applied to various fields, ranging from text classification \cite{kulis2011you, shi2010transfer,wang2011heterogeneous}, spam email and intrusion detection \cite{chattopadhyay2012multisource, gao2008knowledge, gao2009graph, luo2008transfer}, recommendation systems \cite{moreno2012talmud, cao2010transfer, jiang2012social, li2009can, li2009transfer, pan2010transfer, pan2011transfer,raina2007self, zhang2012multi, zhao2013active}, biology and gene expression modeling \cite{ogoe2015knowledge, widmer2012multitask}, to image and video concept classification \cite{kan2014domain, farhadi2007transfer, patel2015visual,shao2014transfer}, human activity recognition \cite{electronics10192412,app11167660,9447028}. While these fields are vastly different, they all benefit from the core functionalities of transfer learning, in applying the knowledge gained under controlled settings or similar domains, to new areas that may otherwise lack this knowledge.

\subsection{Challenges}
Despite many successes in the area of transfer learning, some challenges still remain. This section discusses current challenges and possible improvements.

\textbf{Negative Transfer}
One of the earliest challenges discovered in transfer learning is called negative transfer learning. The term describes when the transfer results in a reduction in performance. One of the reasons could be the interference with previous knowledge\cite{https://doi.org/10.48550/arxiv.2105.05837} or the dissimilarity between the domains\cite{raghu2019transfusion,https://doi.org/10.48550/arxiv.2006.06882} could be one of the reasons. There might be some cases where the transfer does not degrade, but doesn't make full use of its potential to obtain a representative feature. In \cite{https://doi.org/10.48550/arxiv.2105.05837} has been shown that contrastive pre-training on the same domain may be more effective than attempting to transfer knowledge from another domain. Similarly, in \cite{https://doi.org/10.48550/arxiv.1905.07553} the study was conducted to explore which tasks will gain from sharing knowledge and which will suffer from negative transfer and should be learned in a separate model. In \cite{Wang_2019_CVPR} the authors proposes a formal definition of negative transfer and analyzes three key aspects, as well as a model for filtering out unrelated source data.

\textbf{Measuring Knowledge Gain}
The concept of transfer learning enables remarkable gains in learning new tasks. However, it's difficult to quantify how much knowledge is transferred. A mechanism for quantifying transfer in transfer learning is essential for understanding the quality of transfer and its viability. In addition to the available evaluation metrics, we need to assess the generalizability/robustness of the models, especially in situations where class sets are different between problems\cite{sousa2014transfer}. There was an attempt in\cite{glorot2011domain,moon2017completely,Wang_2019_CVPR} to formulate the problem so that transfer learning related gains could be quantified.

\textbf{Scalability and Interpretability}
Although many works demonstrate the ability of tasks to be transferred and their effectiveness, there is no guideline on how and what should be transferred. It has been shown that transfer learning can be effective only when there is a direct relationship between source and target; however, there have been many instances where transfer learning has failed despite the assumption of reletivity. Furthermore, as pre-trained models are becoming more widespread, with millions or billions of papameters, it would not be feasible to try all of the available methods to see which transfer could be helpful. Moreover, this requires a tremendous amount of computation, resulting in a large carbon footprint\cite{https://doi.org/10.48550/arxiv.1906.02243,https://doi.org/10.48550/arxiv.2007.03051
}. It is critical that models are interpretable not only for their task, but also in terms of their ability to be transferred to other tasks. This work\cite{8679150} defines the interpretable features that will be able to explain the relationship between the source and target domain in a transfer learning task.

\textbf{Cross-modal Transfer}
In general, transfer learning is used when the source and target domains have the same modalities or input sizes. However, in many scenarios, this assumption could present a problem in adopting knowledge. Our ability to transfer knowledge from different modalities is crucial, since many tasks in our daily lives require information from multiple sources (perception and text or speech). One of the most recent studies, Bert\cite{https://doi.org/10.48550/arxiv.2103.00020} and ViLBert\cite{https://doi.org/10.48550/arxiv.1908.02265}, attempts to transfer knowledge between text and image data. Additionally, we should be able to transfer knowledge regardless of the difference between input sizes in the source and target domain. An example could be transferring knowledge from 2D to 3D datasets\cite{shan20183,8784845}.

\textbf{How to build Transferable models}
The development of neural networks and deep learning models often requires significant architecture engineering. In addition, these models are engineered to outperform the existing models on the target dataset. As a result of the performance gain, the model's ability to generalize is usually degraded. We should be able to build models that enable transferability and reduce the dataset bias. As shown in \cite{pmlr-v37-long15}, deep features eventually transition from general to specific along the network, which make the feature transferability drops significantly in higher layers. Works in \cite{Zoph_2018_CVPR,Lee_2019_ICCV,NEURIPS2019_fd2c5e46,moon2017completely,wang2019transferable,pmlr-v37-long15} try to build the model with the focus of the transferability across domains.

\section{Neural Radiance Fields}
\label{sec:nerf}
Several contributions in computer graphics have had a major impact on deep learning techniques to represent scenes and shapes with neural networks. A particular aim of the computer vision community is to represent objects and scenes in a photo-realistic manner using novel views. It enables a wide range of applications including cinema-graph\cite{10.1111/cgf.12147,https://doi.org/10.48550/arxiv.1801.09042}, video enhancement \cite{Zhang_2021_CVPR,Son_2021_CVPR}, virtual reality \cite{Gao_2021_ICCV}, video stabilization\cite{Xu_2021_ICCV,Liu_2021_ICCV} and to name a few. 

The task involves the collection of multiple images from different viewpoints of a real world scene, and the objective is to generate a photo-realistic image of such a novel view in the same scene. Many advancements have been made, one of the most common is to predict a 3D discrete volume representation using a neural network/cite[Lombardi:2019] and then render novel views using this representation. Usually these models take in the images and pass them through a 3D CNN model\cite{6165309}, then the model outputs the RGBA 3D volume \cite{https://doi.org/10.48550/arxiv.2004.11364,https://doi.org/10.48550/arxiv.1906.07316,https://doi.org/10.48550/arxiv.1905.00889}. Even though these models are very effective for rendering, they don't scale since each scene requires a lot of storage. A new approach to scene representation has emerged in recent years, in which the neural network represents the scene itself. In this case, the model takes in the $X,Y,Z$ location and outputs the shape representation\cite{https://doi.org/10.48550/arxiv.1901.05103, https://doi.org/10.48550/arxiv.1812.03828, https://doi.org/10.48550/arxiv.1906.01618, https://doi.org/10.48550/arxiv.1912.07372}. The output of these models could be distance to the surface \cite{https://doi.org/10.48550/arxiv.1901.05103}, occupancy\cite{https://doi.org/10.48550/arxiv.1812.03828}, or a combination of color, and distance\cite{https://doi.org/10.48550/arxiv.1906.01618, https://doi.org/10.48550/arxiv.1912.07372}. As the shape itself is a neural network model, it is difficult to optimize it for different renderings. However, the key advantage is the shapes are compressed by the neural network which makes it very efficient in terms of memory. Nerf \cite{https://doi.org/10.48550/arxiv.2003.08934} combines these ideas into a single architecture. Given the spatial location $X,Y,Z$ and viewing direction $\theta,\phi$ , a simple fully connected outputs the color $r,g,b$ and opacity $\sigma$ of the specified input location and direction. 

A very high level explanation of Nerfs could be think of as function that can map the the 3D location($x$) and ray direction($d$) to the color($r,g,b$) and volume density($\sigma$).

$$
\mathrm{F}(\underline{\mathbf{x}}, \underline{\mathbf{d}})=(\mathrm{r}, \mathrm{g}, \mathrm{b}, \boldsymbol{\sigma})
$$

During the training stage, given a set of image from different views(well-known camera poses) an MLP is trained to optimize it weights. 

In order to generate a realistic photo, we have to hypothetically place the camera(having the position) and point it to a specific direction. 

Consider from the camera we shoot a ray and we want to sample from the NeRF along the way. There might be a lot of free space, but eventually the ray should collide with the surface of the object. The summation along the ray should represent the pixel's color at the specific location and viewing direction. In other words, the pixel value in image space is the weighted combination of these output values as below.

\begin{equation}
C \approx \sum_{i=1}^N T_i \alpha_i c_i
\end{equation}

Where $T_i$, can be think of as weights, is the accumulated product of all of the values behind it:

\begin{equation}
T_i=\prod_{j=1}^{i-1}\left(1-\alpha_j\right)
\end{equation}

where $\alpha_i$ is:

\begin{equation}
\alpha_i=1-e^{-\sigma_i \delta t_i}
\end{equation}

In the end, we can put all the pixels together to generate the image. The whole process, including the ray shooting, is fully differentiable, and can be trained using the total squared error:

\begin{equation}
\min _\theta \sum_i\left\|\operatorname{render}_i\left(F_\theta\right)-I_i\right\|^2
\end{equation}

Where $i$ representing the ray and the loss minimizing the error between the rendered value from the network, $F_\theta$, and the $I$ is the actual pixel value. 

\section{Challenges}
In spite of the many improvements and astounding quality of rendering, the original Nerf paper left out many aspects. 

One of the main assumptions in the original Nerf paper was static scenes. For many applications, including AR/VR, video game renderings, objects in the scenes are not static. The ability to render objects with respect to time along with the novel views are essential in my applications. There are some works attempting to solve the problem and change the original formulation for dynamic scenes and non-rigid objects\cite{Park20arxiv_nerfies,Pumarola20arxiv_D_NeRF,Rebain20arxiv_derf,Jiang2022lord}.

The other limitation is slow training and rendering. During the training phase, the model needs to qeury every pixel in the image. That results about 150 to 200 million queries for a one megapixel image\cite{https://doi.org/10.48550/arxiv.2003.08934}, also, inference takes around 30 sec/frame. In order to solve the training issue, \cite{kangle2021dsnerf} proposes to use the depth data, which makes the network to need less number of views during training. Other network properties and optimizations can be change to speed-up the training issue\cite{https://doi.org/10.48550/arxiv.2111.11215,https://doi.org/10.48550/arxiv.2106.14942}. Inference also needs to be real-time for many rendering applications. Many works try to address the issue in different aspects; changing the scene representation to voxel base \cite{wu2021diver,Liu20neurips_sparse_nerf}, separate models for foreground and background \cite{kaizhang2020} or other network improvements \cite{https://doi.org/10.48550/arxiv.2103.10380,Rebain20arxiv_derf,Wang_2022_CVPR,https://doi.org/10.48550/arxiv.2206.00878}. 

A key feature of the representation for real-world scenarios is the ability to generalize across many cases. In contrast, the original Nerf trained an MLP for every scene. Every time a new scene is added, the MLP should be retrained from scratch. Several works have explored the possibility of generalizing and sharing the representation across multiple categories or at least within the category\cite{Schwarz20neurips_graf,Yu20arxiv_pixelNeRF,jang2021codenerf,SRF}.

For the scope to be widened to other possible applications, we need control over the renderings in different scenarios. The control over the camera position and direction was examined in the original Nerf paper. Some works attempted to control, edit, and condition it in terms of materials\cite{baatz2021nerf,https://doi.org/10.48550/arxiv.2111.15246}, color \cite{liu2021editing,jang2021codenerf}, object placement \cite{yang2021objectnerf,https://doi.org/10.48550/arxiv.2011.12100,yang2021objectnerf}, facial attributes\cite{kania2022conerf,https://doi.org/10.48550/arxiv.2111.15490} or text-guided editing \cite{wang2021clip}.

\section{The Challenges of Representation Learning}
\label{sec:challenges}

Numerous challenges shall be addressed while learning representations from the data. The following section will provide a brief discussion on the most prominent challenges faced in the deep representation learning.

\textbf{Interpretability:} 
There is a fine distinction between explainability and interpretability of a system. An explanation can be defined as any piece of information that helps the user understand the model's behavior and the process that it goes through to make the decision. Explanations can give insights on the role of each attribute in the overall performance of the system, or rules that determine the expected outcome, i.e., when a condition is met\cite{Ras2022ExplainableUninitiated}. Interpretability, however, is considered as a human's ability to predict what the model result would be, based on the decision flow that the model follows \cite{kim2016examples}. A highly interpretable ML model is an easily comprehensible one, but the deep neural networks miss this aspect. Despite the promising performance of deep neural networks in various applications, the inherent lack of transparency in the process by which a deep neural network provides an output is still a major challenge. This black-box nature may render them useless in several applications, such as in situations where high degree of safety \cite{ault2019_interpretable_traffic}, security \cite{ross2017_security_interpret}, fairness and ethicality \cite{Binns2018FairnessIM}, or reliability \cite{kustner_automated_2018,khoshnevisan_2018,Retain_2016,chen_deep_2019,baghaei_sepsis2019} are critical. 

Therefore, design and implementation of problem-specific methods of interpretability and explainability is necessary\cite{alibak2022simulation}. Although the conventional methods of learning from data, such as decision trees, linear models, or self-organization maps \cite{pourkia_2019}, may provide visual explainability \cite{lipton_mythos_2017}, deep neural network require post-hoc methods of interpretation. From a trained model, the underlying representation of the input data, may be extracted and presented in understandable formats for the end-users. Examples of post-hoc approaches are \cite{lipton_mythos_2017}: sentences generated as explanations \cite{lipton_mythos_2017}, visualizations \cite{maaten_visualizing_2008}, explanation by examples \cite{doshivelez2014graphsparse}. Granted that, the post-hoc approaches provide another representation of the captured features, they do not directly reveal the exact causal connections and correlations at the model parameters level \cite{lipton_mythos_2017}. Nonetheless, it increases the reliability of the deep models.

\textbf{Scalability:}
Scalability is an essential and challenging aspect of many representation learning models, partly because getting models to maintain the quality and scale up to real-world applications relies on several different factors, including high-performance computing, optimized workloads distribution,  managing a large distributed infrastructure, and Generalization of the algorithm\cite{10.1145/3363554,10.1145/3070607.3070612, osti_1846568 }. \cite{10.1145/2487788.2488042} has classified big-data machine learning approaches based on distributed or non-distributed fashion. In general, the scalability of representation learning models faces multiple dimensions and significant technical challenges:
  1) availability of large amount of data 2) scaling the model size 3) scaling the number of models and/or computing machines 4)computing resources that can support the computational demands \cite{10.1145/3070607.3070612, osti_1846568,gaonkar2020ethical}\\
The huge amount of data can be accessed from a variety of sources, including internet clicks, user-generated content, business transactions, social media, sensor networks, etc\cite{10.2307/43589260} . Despite the growing pervasiveness level of big data, there are still challenges to accessing a high-quality training set. Data sharing agreements, violation of privacy\cite{Beaulieu-Jones159756,8677282}, noise problem\cite{GUPTA2019466,app11094132}, poor data quality(fit for purpose) \cite{9417095}, imbalance of data\cite{10.1145/3343440}, and lack of annotated datasets are number of challenges businesses face seeking raw data. Oversampling, undersampling, dynamic sampling \cite{johnson2019survey} for imbalanced data, Surrogate Loss, Data Cleaning, finding distribution in solving the problem of learning from noisy labels for noisy data sets, and active learning\cite{series/synthesis/2012Settles} for lack of annotated data are a number of methods have been proposed to alleviate these problems.\\
Model scalability is one of the other concerns in which tasks may exhibit very high dimensionality. To efficiently handle this requirement, different approaches are proposed that cover the last two significant technical challenges mentioned earlier: using multiple machines in a cluster to improve the computing power (scaling out)\cite{10.5555/3042817.3043086} or using more powerful graphics processing units. Another crucial challenge is managing a large distributed infrastructure that hosts several deep learning models trained with a large amount of data. Over the last decade, there have been several types of research done in the area of high-performance computing to alleviate open research problems in infrastructure and hardware, Parallelization Methods, Optimizations for Data Parallelism, Scheduling and Elasticity, Data Management \cite{10.1145/3363554} \cite{10.1145/3003665.3003669,222611,186212,inproceedings,10.5555/1863086.1863099}. While building large clusters of computing nodes may face several problems, such as communication bottlenecks, on the other hand, attempts to accelerate the performance of GPUs capable of implementing energy-efficient DL execution run across several major hurdles\cite{10.5555/1863086.1863099,9842579}.
Though we are able to train extremely large neural networks, they may optimize for a single outcome, and several challenges still remain.


%

\leavevmode \\
\textbf{Security, Robustness, Adversarial Attacks:}
Machine learning is becoming more widely used, resulting in security and reliability concerns. Running these AI workflows for real-world applications may be vulnerable to adversarial attacks. AI models are developed under carefully controlled conditions for optimal performance. However, these conditions are rarely maintained in real-world scenarios. These changes could be both incidental or intentional adversity, both could result in a wrong prediction. Efficacy in detecting and detecting adversarial threats is referred to as adversarial robustness. A major challenge in robustness is the non-interpretability of many advanced models' representations. In \cite{zhang2019interpreting} the authors show that there's a positive connections between model interpretability and adversarial robustness. In some cases\cite{https://doi.org/10.48550/arxiv.1708.08296,Fukui_2019_CVPR,Dwivedi_2022_CVPR}, researchers attempt to interpret the results, but they usually pick examples and show the correlation between the representations and semantic concepts. However, such a relationship may not exist in general\cite{Thrush_2022_CVPR,https://doi.org/10.48550/arxiv.2104.06644}. The discontinuity of the representation first introduced in \cite{https://doi.org/10.48550/arxiv.1312.6199}, where deep neural networks can be misclassified by adding imperceptible, non-random noise to inputs. For a more detailed discussion of different types of attacks, readers can refer to \cite{8294186,xu2020adversarial}. It is worth mentioning that Research on adversarial perturbations and attack techniques is primarily carried out in image classification\cite{machado2021adversarial,dong2020benchmarking,https://doi.org/10.48550/arxiv.1711.00117}, the same behavior is also seen in natural language processing\cite{zhang2020adversarial,wallace2020interpreting}, speech recognition\cite{qin2019imperceptible,alzantot2018did,carlini2018audio}, and time-series analysis\cite{karim2020adversarial,fawaz2019adversarial}. For the systems based on biometrics verification\cite{8621025,https://doi.org/10.48550/arxiv.2105.06625,fei2020adversarial}, an adversarial attack could compromise its security. The use of biometrics in establishing a person's identity has become increasingly common in legal and administrative tasks\cite{jain2004introduction}. The goal of representation learning is to find a (non-linear) representation of features $\mathrm f: \mathcal{X} \rightarrow \mathcal{Z}$ fro from input space $ \mathcal{X}$ to feature space $\mathcal{Z}$ so that $f$ retains relevant information regarding the target task $\mathcal{Y}$ while hiding sensitive attributes\cite{NEURIPS2020_6b8b8e3b}. Despite all the proposed defenses, deep learning algorithms still remain vulnerable to security attacks, as proposed defenses are only able to defend against the attacks they were designed to defend against\cite{kang2019transfer}. In addition to the lack of universally robust algorithms, there is no unified metric by which to evaluate the robustness and resilience of the algorithms.



\section{Conclusion}
\label{sec:conclusion}
Since many of the state-of-the-art architectures rely on variants of deep neural networks to achieve competitive performances, the methods of representing data, can be considered as the building blocks of the proposed methods. It is therefore essential to have a thorough understanding of the major methods for learning representations. In this article, we reviewed the fundamentals and basics of various approaches to learning representations from data. Throughout this presentation, we sought to explain each topic concisely so that people who are new to the field or work on interdisciplinary projects-- possibly with backgrounds other than computer science-- can gain an understanding of the topics discussed. In addition, we provide detailed references and applications of each concept so that the interested readers can deepen their understanding of the concepts.


\bibliographystyle{IEEEtran}
\bibliography{refs}

\end{document}